\documentclass{article}

\usepackage[preprint]{neurips_2026}

\usepackage{enumitem}

\usepackage{enumitem}
\usepackage{pifont}
\usepackage{balance}
\usepackage{multicol}
\usepackage{multirow}
\usepackage{xcolor}
\usepackage{colortbl}
\definecolor{mygray}{gray}{.9}
\usepackage{caption}
\usepackage{makecell}
\usepackage{algorithm}
\usepackage{algorithmic}
\usepackage{graphicx}

\usepackage[utf8]{inputenc} 
\usepackage[T1]{fontenc}    
\definecolor{cvprblue}{rgb}{0.21,0.49,0.74}
\usepackage[pagebackref,breaklinks,colorlinks,allcolors=cvprblue]{hyperref}
\usepackage{url}            
\usepackage{booktabs}       
\usepackage{amsfonts}       
\usepackage{nicefrac}       
\usepackage{microtype}      
\usepackage{xcolor}         
\usepackage{titletoc}
\usepackage{amsmath}        
\usepackage{wrapfig}        

\usepackage{placeins}
\usepackage[capitalize]{cleveref}

\usepackage[utf8]{inputenc} 
\usepackage[T1]{fontenc}    
\usepackage{hyperref}       
\usepackage{url}            
\usepackage{booktabs}       
\usepackage{amsfonts}       
\usepackage{nicefrac}       
\usepackage{microtype}      
\usepackage{xcolor}         
\usepackage{amsmath,amssymb}

\usepackage[utf8]{inputenc} 
\usepackage[T1]{fontenc}    
\usepackage{hyperref}       
\usepackage{url}            
\usepackage{booktabs}       
\usepackage{amsfonts}       
\usepackage{nicefrac}       
\usepackage{microtype}      
\usepackage{xcolor}         

\title{$\text{DataClaw}_0$: Agentic Tailoring Multimodal Data from Raw Streams}

\author{
Cong Wan\textsuperscript{1}$^\ast$,
Zeyu Guo\textsuperscript{1}$^\ast$,
Zijian Cai\textsuperscript{2}$^\ast$,
Jiangyang Li\textsuperscript{1}$^\ast$,
SongLin Dong\textsuperscript{3},
Lin Peng\textsuperscript{1}
\and
Xiangyang Luo\textsuperscript{4},
Zhiheng Ma\textsuperscript{3},
Yihong Gong\textsuperscript{1},
\\[0.8em]
\textsuperscript{1}Xi'an Jiaotong University\\
\textsuperscript{2}University of Chinese Academy of Sciences\\
\textsuperscript{3}Shenzhen University of Advanced Technology\\
\textsuperscript{4}Tsinghua University
}

\begin{document}

\maketitle

\begin{abstract}
Raw multimodal streams -- tutorial videos, embodied trajectories, GUI interaction logs -- are abundant but noisy, redundant, and unaligned with any particular training objective. Turning them into supervision today means either brittle heuristics or repeatedly querying a proprietary vision-language model, a cost that recurs with every new sample. We ask whether this conversion can instead be \emph{learned once and reused}, and formalise intent-conditioned \textbf{Data Tailoring}: given a raw stream and a high-level intent, a model must return schema-aligned, evidence-grounded training instances. Training $\text{DataClaw}_0$ at 4B, 9B and 27B, we find that whether five heterogeneous domains should share one model depends on capacity. A jointly trained model is \emph{worse} than per-domain experts at the two smaller scales and \emph{better} at the largest, placing the crossover near 18B parameters. Matched-data comparisons, in which the joint model sees exactly the same data per domain as that domain's expert, attribute the reversal to cross-domain transfer rather than to data volume: it gains most where a domain is data-poor, and recovers 59\% of in-domain performance on domains withheld from training entirely. Downstream post-training reproduces this ordering on GUI navigation, action video generation and spatio-temporal VQA, and the joint configuration is also the cheaper to deploy, serving one model instead of five.
Github: \url{https://github.com/vancyland/DataClaw0}
\end{abstract}

\section{Introduction}

The rapid evolution of multimodal foundation models is increasingly bottlenecked by a critical scarcity of high-quality training data~\cite{qwen2.5vl, qwen3vl,llava,gpt4o,gpt4,gemini,gemini2.5,gpt5,claude,minimax,wan2025,seedance2}. While the physical and digital worlds generate an inexhaustible supply of raw multimodal streams—such as hours-long tutorial videos, embodied agent trajectories, and complex Web and GUI operation logs~\cite{grauman2022ego4d,savva2019habitat,dai2017scannet,androidinthewild,androidworld,mind2web}—harvesting this resource presents a formidable challenge. The core obstacle is extreme ``data entropy''. Unlike curated image-text pairs, raw multimodal streams are inherently noisy, highly redundant, and weakly structured. They contain dense physical dynamics, procedural knowledge, and implicit decision logic~\cite{grauman2022ego4d,Ego-exo4d,miech2019howto100m,savva2019habitat,osworld,mind2web}, but lack the explicit supervision signals required for efficient knowledge acquisition or high-quality model post-training~\cite{textbooks,lima,sharegpt4v,llavaonevision}. Consequently, efficiently distilling continuous, high-entropy streams into structured, high-density knowledge has become the most pressing imperative in multimodal data engineering.

Existing data processing pipelines remain largely passive, relying on heuristic sampling, coarse captioning, or directly prompting general Vision-Language Models (VLMs) to generate captions and question-answer pairs~\cite{sharegpt4v,gao2024sphinx,chen2023minigptv2}. While effective for short and curated inputs, these methods struggle with long, noisy streams that require temporal reasoning, spatio-temporal consistency, and implicit physical understanding~\cite{vlm2bench,vsibench,mmsibench,disheng2024thinking,mico,li2025viewspatial}. Direct annotation therefore often yields hallucinated, fragmented, or low-density outputs, leaving much of the latent value in raw multimodal data untapped~\cite{hallusionbench,evaluating}. Equally important, prompting a proprietary VLM is a \emph{recurring} cost: every additional sample is billed again, so the expense of data production scales linearly with the amount of data one needs. This motivates us to recast high-quality data production as a learnable capability, termed \textbf{Data Tailoring}: given a high-level user intent or downstream training objective, a tailoring model filters redundant information, identifies task-critical evidence, and reorganizes it into dense, verifiable, and application-specific supervision. Unlike general data curation or synthetic instruction generation~\cite{data-centric-survey,dataperf,selfinstruct,wizardlm,orca,lima,alpagasus}, data tailoring focuses on intent-conditioned entropy reduction over continuous multimodal streams. This paper asks whether such a capability can be formally defined, rigorously evaluated, and learned once by a compact open-source VLM so that it can subsequently be reused at low marginal cost.


\begin{figure}[t]
  \centering
  \includegraphics[width=1.0\linewidth]{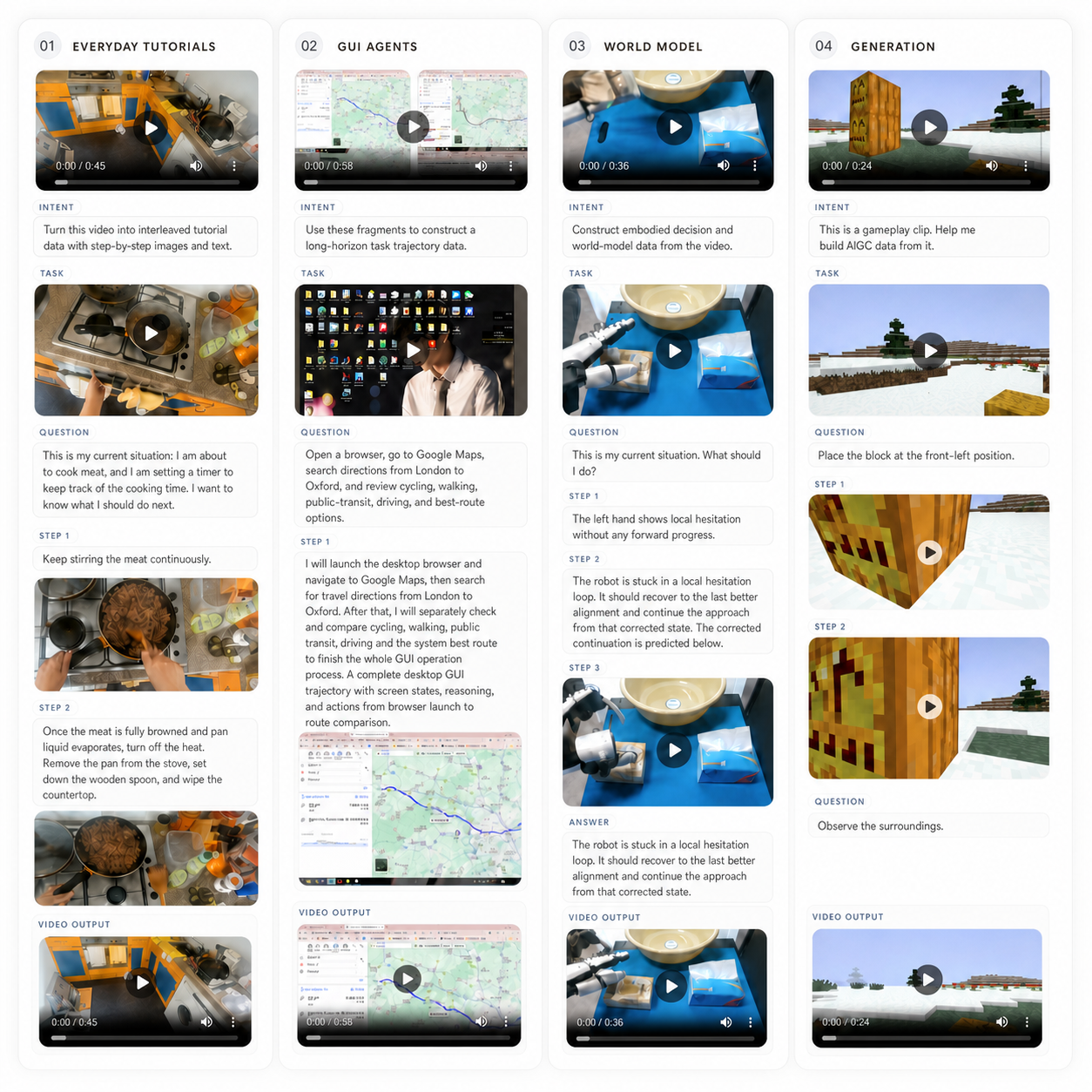}
  \caption{
The figure shows representative tailoring cases. In each panel, the top clips denote the raw inputs, followed by a user \textit{Intent} that specifies a construction goal. $\text{DataClaw}_0$ then selects task-relevant visual evidence under \textit{Task}, formulates a corresponding \textit{Question}, and decomposes the solution into intermediate \textit{Steps} with aligned images and textual reasoning/action descriptions. The bottom part gives the final structured outcome, including an \textit{Answer} when applicable and a \textit{Video Output} that preserves or reconstructs the tailored visual sequence. 
  }
  \label{fig:intro_cases}
\end{figure}

To answer this, we must first overcome the data paradox: training a model to refine data requires high-quality refinement data. We propose a scalable, automated two-stage construction pipeline. Our first stage extracts deterministic Factual Anchors using lightweight domain experts, metadata parsers, and heuristic rules. These anchors provide reliable low-level grounding, such as object states, temporal boundaries, OCR text, and GUI interaction events~\cite{savva2019habitat,dai2017scannet,vlmad,spatialvlm,robospatial}. In the second stage, strong VLMs perform long-range logical chaining over these discrete anchors, injecting multi-dimensional reasoning traces inspired by multimodal chain-of-thought~\cite{zhang2023multimodalcot,chen2024measuring,qian2024visual}. This bottom-up extraction and top-down synthesis strategy yields a massive, cross-domain refinement dataset spanning five representative arenas: daily life, education, embodied intelligence, world models, and GUI agents~\cite{coin,omniworld,worldmem,epic-kitchens,wan2024grid,unireal,stepxedit,qwenimage,openvla,pi0,gr3,worldvla}.

Leveraging this dataset, we introduce \textbf{$\text{DataClaw}_0$}, a framework for intent-conditioned multimodal data tailoring, illustrated in Fig.~\ref{fig:intro_cases}. At its core, $\text{DataClaw}_0$ optimizes Qwen3.5-27B to transform high-entropy streams into intent-aligned, customized structured outputs, and we repeat the entire recipe at 4B and 9B so that conclusions about architecture can be separated from conclusions about capacity. Moving beyond standard Supervised Fine-Tuning (SFT), $\text{DataClaw}_0$ employs Group Relative Policy Optimization (GRPO)\cite{guo2025deepseekmathgrpo} to directly optimize for tailoring quality and intent adherence. The reinforcement learning phase utilizes multi-dimensional reward signals that measure intent satisfaction, information density, factual consistency, and structural correctness, drawing inspiration from recent reasoning-oriented alignment techniques\cite{rafailov2024dpo,sun2024llava-rlhf,wang2024reasoning,li2024enhancing}. We train two configurations at every scale, not as alternative products but as the two sides of the question this paper asks: $\text{DataClaw}_0$-O, one model trained jointly on all five domains, and $\text{DataClaw}_0$-E, a set of decoupled domain-specific experts with a router. Which of the two is preferable is an empirical matter, and we find that the answer changes with backbone capacity.

To systematically evaluate this capability, we construct $\text{DataClaw}_0$-val (including an Intent subset for vague-intent concretization), a benchmark dedicated to structured data refinement.
On this benchmark our 27B model surpasses substantially larger proprietary annotators on all three axes, whereas the 4B and 9B variants do not. More interestingly, whether the five domains should share one model depends on capacity. Across 4B, 9B and 27B backbones, sharding into per-domain experts is better at the two smaller scales and worse at the largest, and matched-data comparisons show the reversal is driven by positive transfer between domains rather than by the joint model seeing more data. Because a benchmark of our own design cannot by itself establish practical value, we additionally introduce a Targeted Refinement evaluation across video generation, real-world VQA, and GUI navigation~\cite{grauman2022ego4d,remot,deng2009imagenet}, and treat it as the primary criterion. Under a fixed data budget, models trained on $\text{DataClaw}_0$-tailored subsets outperform those trained on data produced by the downstream model itself and by a proprietary annotator, and the joint-versus-sharded ordering we observe on the benchmark reproduces on all three downstream tasks. Finally, we report an amortised cost analysis (Appendix~\ref{app:cost}): distilling the tailoring capability incurs a one-off cost, but the marginal cost of each subsequent sample is that of local inference. Two design axes must be separated here, because quality alone conflates them. At a fixed backbone, the joint configuration is also the cheaper one to deploy, since it serves one model rather than five. Across backbones the trade runs the other way: the 9B variant costs \$0.009 per sample against \$0.038 for the teacher API and breaks even after about 80K samples, whereas the more accurate 27B model costs \$0.027 and needs about 324K. We report the second trade rather than only the first, and state in Appendix~\ref{app:cost} the accelerator prices at which neither pays off. 

In summary, our core contributions are: 
(1) We formulate multimodal data processing as a learnable capability, \textbf{Data Tailoring}, and propose a two-stage pipeline of deterministic anchor extraction and generative semantic synthesis that yields a cross-domain training corpus together with $\text{DataClaw}_0$-val, a benchmark for structured data refinement.
(2) We propose $\text{DataClaw}_0$, a tailoring framework combining SFT with GRPO, and study both Omni and Expert deployment paradigms, and use it to establish our main finding: cross-domain positive transfer in data tailoring emerges above a capacity threshold, which we locate near 18B parameters by training at 4B, 9B and 27B. Expert routing is therefore the better design below the threshold and joint training above it. We support this with matched-data, cross-domain and held-out-domain experiments rather than with aggregate scores alone.
(3) We introduce a Targeted Refinement setting that uses downstream post-training on GUI navigation, action video generation, and spatio-temporal VQA as the primary criterion, and pair it with an amortised cost analysis reporting the break-even volume at which self-hosted tailoring becomes cheaper than API annotation, together with the configurations and accelerator prices under which it does not.
\section{Related Work}

\paragraph{Multimodal Large Language Models.}
Large language models (LLMs) have demonstrated strong reasoning, instruction-following, and generalization capabilities in text-only scenarios~\cite{gpt4,videollama,yang2025qwen3}. By integrating visual encoders, multimodal projectors, and language backbones, multimodal large language models (MLLMs) extend these capabilities to visual perception, grounding, and multimodal reasoning~\cite{llava,chen2023minigptv2,qwen2.5vl,chen2024internvl,gpt4o,gemini2.5}. Recent MLLMs have rapidly evolved from image-level assistants to general-purpose multimodal systems, covering high-resolution perception, OCR, document understanding, multi-image reasoning, video understanding, spatial reasoning, and long-context multimodal interaction~\cite{spatialvlm,vlm4d,vlmad,vlm2bench,vsibench,mmsibench}. Beyond generic perception and reasoning, MLLMs are increasingly adapted to downstream scenarios in both digital and physical environments, including GUI interaction, web/mobile automation, world model, and vision-language-action modeling~\cite{osworld,mind2web,savva2019habitat,dai2017scannet,openvla,pi0,gr3,worldvla,wan2024grid,li2026trajectory,luo2025canonswap,retrieve,prosr}. These applications require models to interpret complex observations, localize task-critical entities, understand temporal or procedural dependencies, and sometimes transform multimodal states into executable actions. However, the continued scaling of MLLMs is increasingly constrained by data. High-quality multimodal supervision is costly to obtain, while raw multimodal streams, such as tutorial videos, embodied trajectories, 3D scans, and GUI operation logs, are usually noisy, redundant, weakly structured, and poorly aligned with downstream training objectives~\cite{grauman2022ego4d,savva2019habitat,dai2017scannet,osworld,mind2web}. Existing passive annotation pipelines, including heuristic sampling, coarse captioning, and direct VLM-based question-answer generation, have enabled early multimodal instruction tuning~\cite{sharegpt4v,gao2024sphinx,chen2023minigptv2}, but they struggle to extract dense and reliable supervision from long, high-entropy streams, especially under temporal reasoning, spatial consistency, and hallucination-sensitive settings~\cite{hallusionbench,evaluating,disheng2024thinking,mico,li2025viewspatial}. Therefore, scalable transformation of raw multimodal data into structured, task-aligned supervision has become a central bottleneck for pursuing the ceiling of general intelligence.

\paragraph{Agentic AI.}
Recent LLM agents extend language models from passive response generators to goal-driven systems that can plan, use tools, interact with environments, execute multi-step workflows, and refine intermediate results through feedback~\cite{yao2022react,schick2023toolformer,shen2023hugginggpt,yao2023tree,shinn2023reflexion,wang2023voyager}. This agentic paradigm has attracted broad attention in both research and open-source communities, with systems and frameworks such as OpenAgents, LangChain, AutoGPT, and OpenClaw enabling language-model agents to orchestrate tools, access external resources, automate user workflows, and operate across heterogeneous digital environments~\cite{xie2023openagents,langchain2023,autogpt2023,openclaw2025}. Beyond task execution, agentic workflows have also been explored for data-centric applications, including synthetic instruction generation, self-improvement, and high-quality reasoning data construction~\cite{selfinstruct,wizardlm,orca,chen2024agentinstruct}. These studies suggest that agents can serve not only as executors, but also as scalable organizers of complex data-generation pipelines: they can decompose ambiguous objectives, call specialized tools, inspect intermediate outputs, correct errors, and verify results. Inspired by this observation, $\text{DataClaw}_0$ formulates MLLM data synthesis as an agentic workflow. Instead of manually annotating or directly captioning raw streams, $\text{DataClaw}_0$ organizes multimodal models and tools, into a data-synthesis agent that converts high-entropy multimodal sources into structured, intent-aligned supervision for training models.
\section{Method}

\begin{figure}[t]
  \centering
  \includegraphics[width=0.99\linewidth]{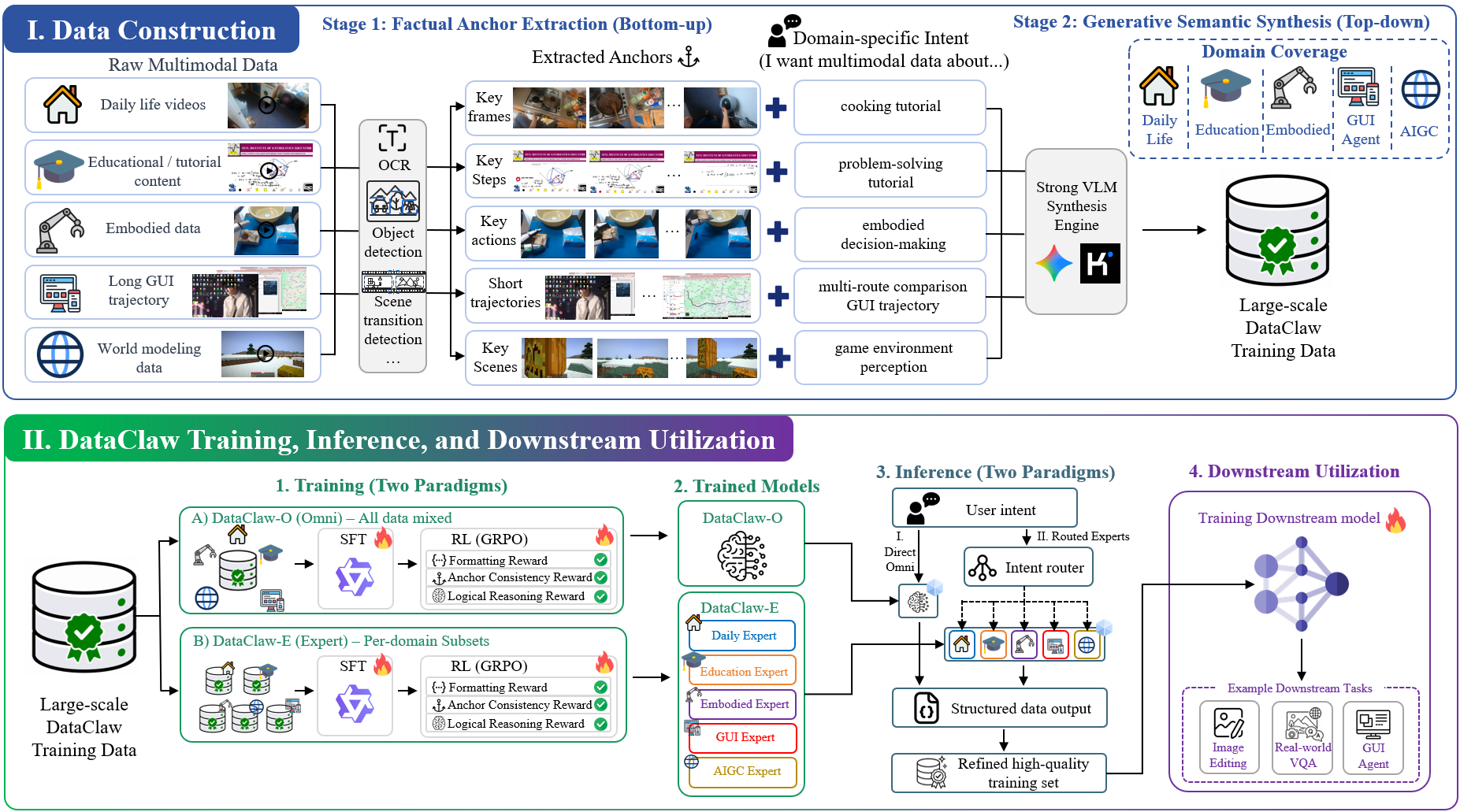}
  \caption{
  \textbf{Overview.}
  The pipeline consists of two parts. First, $\text{DataClaw}_0$ constructs training data by extracting bottom-up factual anchors from raw multimodal data, including key frames, steps, actions, trajectories, and scenes, and then combining them with domain-specific intents for top-down semantic synthesis by a strong VLM. This produces large-scale structured data across daily life, education, embodied, GUI-agent, and AIGC domains. Second, the constructed data is used to train $\text{DataClaw}_0$ models under two paradigms: an omni model trained on mixed-domain data and expert models trained on per-domain subsets. During inference, user intents are handled either directly by the omni model or routed to domain experts, yielding refined structured data for downstream multimodal tasks.
  }
  \label{fig:overview}
\end{figure}

\subsection{Problem Formulation: Intent-Conditioned Data Tailoring}

Traditional multimodal tasks (e.g., Video Captioning or Visual Question Answering) are typically modeled as passive descriptive or question-answering processes. In contrast, $\text{DataClaw}_0$ addresses \textit{Intent-Conditioned Multimodal Data Tailoring}. Here the model must filter, reason over, and reorganize lengthy, noisy raw multimodal streams into high-value structured assets, guided by a specific high-level intent. We use ``tailoring'' rather than ``agentic'' deliberately: the model performs a single conditioned mapping from stream and intent to structured output, and does not plan, call external tools, or iteratively revise its own result at inference time.

We define the input as a raw multimodal data stream:
\begin{equation}
X_{raw} = \{x_1, x_2, \dots, x_T\}
\end{equation}
where \(x_t\) represents the visual frame or multimodal segment at time step \(t\) (e.g., a sequence of frames from a long video or GUI operation screenshots). Simultaneously, an intent instruction \(I\) is provided to represent the user's high-level objective or the downstream task requirement.

The goal of the tailoring agent is to generate customized, structured knowledge assets:
\begin{equation}
Y_{struct} = \{y_1, y_2, \dots, y_L\}
\end{equation}
Unlike free-form text, \(Y_{struct}\) must strictly conform to a predefined structural schema \(\Phi\) (e.g., a specific JSON format, Markdown syntax, or action code logic) tailored to the intent \(I\).

Therefore, the optimization objective of the core $\text{DataClaw}_0$ tailoring agent \(F_\theta\) (parameterized by \(\theta\)) is to maximize the conditional generation probability given the raw data stream and intent, constrained by the structural schema:
\begin{equation}
\theta^* = \arg\max\limits_\theta \sum\limits_{(X_{raw},I,Y_{struct})\in D} \log P(Y_{struct}\mid X_{raw},I;\theta)\cdot \mathbb{I}(Y_{struct}\in\Phi)
\end{equation}
where \(D\) is the training dataset, and \(\mathbb{I}(\cdot)\) is an indicator function that equals 1 if the generated sequence conforms to the structural schema \(\Phi\), and 0 otherwise.

This formulation requires two core capabilities: (1) \textit{Information Filtering and Focusing}, which involves eliminating redundant background noise from \(X_{raw}\) based on \(I\) (where \(T \gg L\)); and (2) \textit{Structural Reorganization}, ensuring that the generated \(Y_{struct}\) is not only semantically accurate but also strictly adheres to the required formatting.

\subsection{Data Construction Pipeline}

To train the tailoring agent \(F_\theta\), we construct large-scale triplets \((X_{raw}, I, Y_{struct})\) through a two-stage automated pipeline. First, a lightweight expert ensemble \(H\) extracts factual anchors from raw multimodal streams:
\begin{equation}
A = H(X_{raw}) = \{a_k=(t_k,p_k,c_k)\}_{k=1}^{K},
\end{equation}
where each anchor records timestamp, spatial position, and local semantic content. These anchors provide reliable grounding signals and reduce hallucinations in long-sequence annotation.

Second, a strong VLM synthesis engine \(S\) generates structured supervision conditioned on the raw input, extracted anchors, and domain intent:
\begin{equation}
Y_{struct}=S(X_{raw}, A, I_{domain}).
\end{equation}
The resulting corpus covers multiple multimodal domains and serves as the foundation for subsequent SFT and RL training of $\text{DataClaw}_0$. Detailed construction procedures, expert modules, and prompting strategies are provided in Appendix~\ref{app:data_pipeline}.

\paragraph{Concrete instantiation.} We state the specific components used throughout this work, since the identity of the teacher determines how our results should be interpreted. The synthesis engine \(S\) is \emph{Gemini-3.1-Pro}, queried once per training example; it is used only to construct the SFT corpus and is never invoked at tailoring time. The anchor extractor \(H\) is not a single model but a per-domain ensemble of lightweight, deterministic components: shot-boundary detection and ASR for video streams, an open-vocabulary detector and multi-object tracker for object states and trajectories, an OCR module for on-screen text, and direct parsers over recorded GUI event logs and robot proprioception. Anchors from parsers and logs are exact by construction; anchors from perception modules are the only ones subject to model error. The base models optimised in our experiments are \emph{Qwen3.5-27B} (headline results) and \emph{Qwen3.5-9B} (scale ablation).

\subsection{Rule-Driven Reinforcement Learning via GRPO}

After SFT, we further optimize $\text{DataClaw}_0$ with rule-driven GRPO to improve formatting reliability, reduce hallucinations, and strengthen spatio-temporal grounding. Instead of training an additional neural reward model, we use deterministic rewards tailored to structured multimodal data:
\begin{equation}
R(Y)=\lambda_1R_{format}(Y,\Phi)+\lambda_2R_{anchor}(Y,A)+\lambda_3R_{eff}(Y),
\end{equation}
where \(R_{format}\) checks schema compliance, \(R_{anchor}\) measures alignment with extracted factual anchors and trajectories, and \(R_{eff}\) discourages overly verbose reasoning.

Given a group of sampled outputs, GRPO normalizes their rewards within the group to estimate relative advantages:
\begin{equation}
\hat{A}^{(g)}=\frac{R^{(g)}-\mu_R}{\sigma_R}.
\end{equation}
The policy is then updated with a clipped objective and a KL regularizer against the reference model. This SFT-initialized, rule-reward optimization enables $\text{DataClaw}_0$ to produce more valid, grounded, and concise structured outputs. Detailed reward definitions and the full GRPO objective are provided in Appendix~\ref{app:grpo}.

\subsection{Inference and Deployment Paradigms}

$\text{DataClaw}_0$ is deployed as a structured multimodal tailoring agent that maps raw streams \(X_{raw}\) and user intents \(I\) to schema-aligned outputs \(Y_{struct}\). 
Its inference pipeline follows three stages: multimodal ingestion and intent parsing, schema-constrained policy inference, and post-hoc grounding verification with factual anchors \(A\). This design improves output validity and keeps the generated structured data grounded in the input stream.

$\text{DataClaw}_0$ supports two deployment paradigms. \textbf{$\text{DataClaw}_0$-O} uses a unified omni model for flexible cross-domain processing, while \textbf{$\text{DataClaw}_0$-E} uses a domain-decoupled expert architecture, where each request is handled by the corresponding expert according to the target scenario or deployment configuration. The omni setting favors simplicity and generality, whereas the expert setting provides stronger domain specialization and modular scalability.

Detailed inference architecture and deployment mechanisms are provided in Appendix~\ref{app:deployment}.

\subsection{Benchmark Construction and Evaluation}

We construct \textbf{$\text{DataClaw}_0$-val} by diversity-aware sampling 200 high-quality examples, ensuring coverage of diverse multimodal inputs, target schemas, and long-tail cases. We also introduce \textbf{$\text{DataClaw}_0$-Intent}, a fuzzy-intent stress test that evaluates whether the agent can infer underspecified user intents from colloquial, ambiguous, or incomplete requests.
We evaluate outputs with a hierarchical metric tailored to structured multimodal data. The metric first enforces JSON validity and then measures schema-field correctness, textual semantic alignment, and trajectory-shape similarity. Full construction details and metric definitions are provided in Appendix~\ref{app:benchmark}.

\section{Experiments}



\subsection{Experimental Setup}
\textbf{Implementation Details:} We initialise $\text{DataClaw}_0$ from Qwen3.5-27B and additionally train a Qwen3.5-9B variant with an otherwise identical recipe, so that the effect of backbone capacity can be separated from that of the training data. We additionally train a Qwen3.5-4B variant, giving three backbone sizes spanning roughly $7\times$ in parameters, since our central claim concerns how behaviour changes with capacity and a reversal between two points alone would not establish it. At each scale we train two configurations: \textbf{-O}, a single model trained jointly on all five domains, and \textbf{-E}, a set of single-domain experts with a lightweight router. Unqualified references to $\text{DataClaw}_0$ denote 27B-O, the configuration we recommend; all three scales are reported throughout because the ordering between -O and -E depends on capacity. During the SFT phase, we utilize 34K strictly cleaned instruction refinement data for 1 epoch. In the GRPO reinforcement learning phase, we set the format reward weight \(\lambda_{fmt}=0.7\), the physical anchor reward weight \(\lambda_{anc}=1.0\), the reasoning efficiency penalty weight \(\lambda_{eff}=0.3\), and the learning rate to \(4\times10^{-6}\). Reward weights and learning rate are shared across all three scales and were tuned only on the 9B variant, so no per-scale tuning advantages the 27B results. The 4B models are trained on $4\times$ A100 GPUs, the 9B models on $8\times$, and the 27B models on $16\times$ with tensor parallelism. Both backbones are evaluated with the same decoding configuration recommended for Qwen3.5 (thinking mode, \(\text{temperature}=1.0\), \(\text{top-}p=0.95\), \(\text{top-}k=20\)).

\subsection{Main Results: Joint vs. Sharded Tailoring}
\label{subsec:main_results}

\begin{table}[!tb]
\centering
\caption{Comparison of $\text{DataClaw}_0$ against state-of-the-art MLLM models. We evaluate models across five distinct domains and a newly introduced \textbf{Fuzzy} instruction subset. Proprietary baselines are \texttt{claude-sonnet-4.6}, \texttt{gpt-4o-1120-global} and \texttt{gemini-3.1-pro-preview}. \textbf{-O} denotes one model trained jointly on all domains, \textbf{-E} a set of single-domain experts with a router. Best per column in \textbf{bold}. 4B and 27B rows pending the final runs.}
\label{tab:main_results}
\small
\setlength{\tabcolsep}{3pt}

\begin{tabular}{llccccccc}
\toprule
\multirow{2}{*}{\textbf{Model}} & \multirow{2}{*}{\textbf{Metric}} & \multicolumn{7}{c}{\textbf{Domains}} \\
\cmidrule(lr){3-9}
& & \textbf{GUI} & \textbf{Embodied} & \textbf{AIGC} & \textbf{Daily} & \textbf{Education} & \textbf{Fuzzy} & \textbf{Overall} \\
\midrule
\multirow{3}{*}{Gemini-3.1-Pro-Preview} 
& Field      & \textbf{100.00} & \textbf{100.00} & \textbf{100.00} & \textbf{100.00} & \textbf{100.00} & 88.74 & 98.12 \\
& Semantic   & 90.01 & \textbf{89.17} & 75.26 & 54.51 & 54.51 & 79.63 & 73.85 \\
& Sequence   & \textbf{99.67} & 67.97 & 33.14 & 51.48 & \textbf{51.48} & 47.25 & 58.50 \\
\midrule
\multirow{3}{*}{Claude-Sonnet-4.6} 
& Field      & 57.50 & \textbf{100.00} & \textbf{100.00} & \textbf{100.00} & \textbf{100.00} & 76.35 & 88.98 \\
& Semantic   & 50.05 & 84.83 & 74.26 & 54.46 & 54.46 & 65.72 & 63.96 \\
& Sequence   & 54.06 & 50.11 & 33.38 & 41.07 & 41.07 & 36.48 & 42.70 \\
\midrule
\multirow{3}{*}{GPT-4o (1120-global)} 
& Field      & \textbf{100.00} & \textbf{100.00} & \textbf{100.00} & \textbf{100.00} & \textbf{100.00} & 83.61 & 97.27 \\
& Semantic   & 84.81 & 87.55 & 69.38 & 54.58 & 80.21 & 74.39 & 75.15 \\
& Sequence   & 80.69 & 46.33 & \textbf{46.95} & 29.33 & 50.71 & 42.57 & 49.43 \\
\midrule
\multirow{3}{*}{MiniMax-M2.7} 
& Field      & 92.50 & \textbf{100.00} & \textbf{100.00} & 97.50 & 95.00 & 73.29 & 93.05 \\
& Semantic   & 78.93 & 79.28 & 69.07 & 44.38 & 43.52 & 61.85 & 62.84 \\
& Sequence   & 77.86 & 51.84 & 6.83 & 32.89 & 17.54 & 32.16 & 36.52 \\
\midrule
\multirow{3}{*}{Qwen3.6-plus} 
& Field      & 70.00 & \textbf{100.00} & 95.00 & 95.00 & 82.50 & 77.58 & 86.68 \\
& Semantic   & 62.64 & 87.40 & 67.82 & 51.20 & 62.14 & 64.37 & 65.93 \\
& Sequence   & 66.96 & 60.33 & 39.71 & 42.44 & 30.85 & 35.92 & 46.03 \\
\midrule
\multirow{3}{*}{Qwen3.5-9B } 
& Field      & 94.87 & \textbf{100.00} & 95.00 & 87.50 & 90.00 & 70.46 & 89.64 \\
& Semantic   & 72.72 & 77.48 & 65.27 & 45.66 & 43.41 & 58.24 & 60.46 \\
& Sequence   & 72.71 & 59.35 & 3.29 & 27.70 & 24.75 & 29.63 & 36.24 \\
\midrule
\multirow{3}{*}{$\text{DataClaw}_0$-4B-O} 
& Field      & 92.50 & 90.00 & 72.50 & 80.00 & 57.50 & 65.16 & 76.28 \\
& Semantic   & 66.12 & 48.94 & 40.27 & 46.83 & 30.51 & 42.68 & 45.89 \\
& Sequence   & 63.47 & 48.16 & 11.35 & 22.04 & 8.62 & 17.51 & 28.52 \\
\midrule
\multirow{3}{*}{$\text{DataClaw}_0$-4B-E} 
& Field      & \textbf{100.00} & \textbf{100.00} & 90.00 & 95.00 & 92.50 & 75.34 & 92.14 \\
& Semantic   & 81.42 & 74.16 & 62.83 & 41.05 & 66.27 & 69.31 & 65.84 \\
& Sequence   & 84.71 & 58.32 & 9.14 & 31.86 & 12.47 & 35.82 & 38.72 \\
\midrule
\multirow{3}{*}{$\text{DataClaw}_0$-9B-O} 
& Field      & \textbf{100.00} & \textbf{100.00} & 85.00 & 92.50 & 70.00 & 78.42 & 87.65 \\
& Semantic   & 80.01 & 63.37 & 55.70 & \textbf{62.61} & 45.71 & 67.35 & 62.46 \\
& Sequence   & 85.70 & 67.01 & 23.90 & 35.05 & 17.41 & 39.84 & 44.82 \\
\midrule
\multirow{3}{*}{$\text{DataClaw}_0$-9B-E} 
& Field      & \textbf{100.00} & \textbf{100.00} & \textbf{100.00} & \textbf{100.00} & \textbf{100.00} & 85.17 & 97.53 \\
& Semantic   & 89.18 & 82.93 & 75.36 & 49.72 & 76.43 & 76.28 & 74.98 \\
& Sequence   & 96.33 & 71.60 & 15.26 & 42.59 & 19.75 & 50.31 & 49.31 \\
\midrule
\multirow{3}{*}{$\text{DataClaw}_0$-27B-E} 
& Field      & \textbf{100.00} & \textbf{100.00} & 95.00 & \textbf{100.00} & 97.50 & 88.31 & 96.80 \\
& Semantic   & 90.15 & 85.64 & 71.83 & 61.82 & 74.96 & 71.86 & 76.04 \\
& Sequence   & 95.28 & 76.14 & 25.41 & \textbf{53.87} & 29.14 & 44.17 & 54.00 \\
\midrule
\multirow{3}{*}{\textbf{$\text{DataClaw}_0$-27B-O [Ours]}} 
& Field      & \textbf{100.00} & \textbf{100.00} & \textbf{100.00} & \textbf{100.00} & \textbf{100.00} & \textbf{96.42} & \textbf{99.40} \\
& Semantic   & \textbf{93.47} & 88.02 & \textbf{80.36} & 61.24 & \textbf{82.51} & \textbf{81.27} & \textbf{81.14} \\
& Sequence   & 98.15 & \textbf{79.62} & 37.62 & 53.19 & 40.83 & \textbf{63.48} & \textbf{62.15} \\
\bottomrule
\end{tabular}
\end{table}

Table~\ref{tab:main_results} compares $\text{DataClaw}_0$ with leading proprietary and open-source MLLMs across five representative domains and the Fuzzy instruction subset. We evaluate three aspects of structured data synthesis: \textit{Field} for schema completeness, \textit{Semantic} for content correctness, and \textit{Sequence} for ordering and structural consistency.

$\text{DataClaw}_0$-27B-O obtains the best overall score on all three axes: Field \(99.40\) against \(98.12\) for Gemini-3.1-Pro-Preview, Semantic \(81.14\) against \(75.15\) for GPT-4o, and Sequence \(62.15\) against \(58.50\) for Gemini. The margin is widest where the task is least specified: on the Fuzzy subset it reaches \(81.27\) Semantic and \(63.48\) Sequence against Gemini's \(79.63\) and \(47.25\).

\begin{figure}[!tb]
  \centering
  \includegraphics[width=\linewidth]{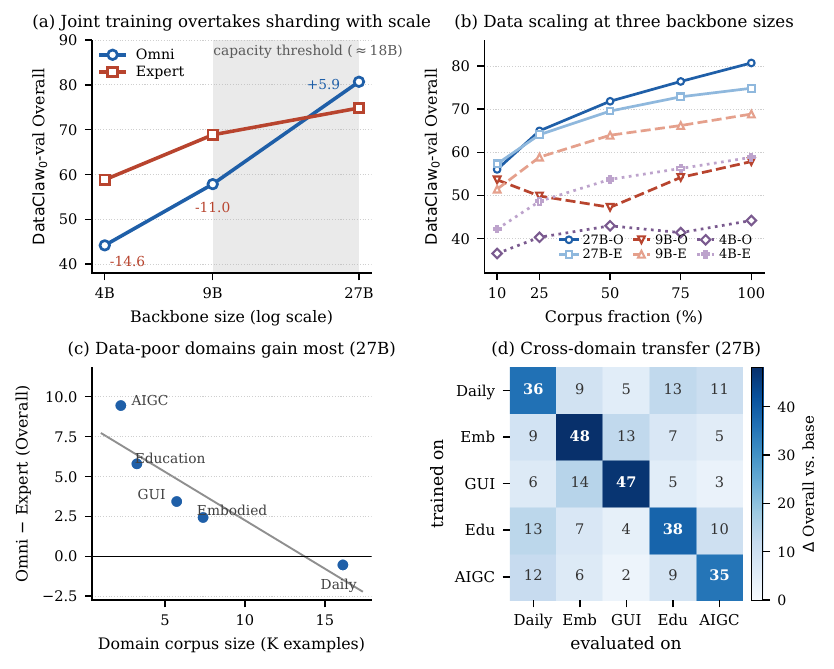}
  \caption{\textbf{(a)} Joint training versus per-domain sharding at three backbone sizes; annotations give the joint-minus-sharded difference. Sharding wins at 4B and 9B and loses at 27B, placing the crossover near 18B parameters (shaded). \textbf{(b)} Data scaling at all three sizes. Both smaller joint curves are non-monotone -- 9B dips at 50\% of the corpus and 4B at 75\% -- whereas the 27B joint curve rises monotonically. Interference shows up as instability, not only as a lower score. \textbf{(c)} The joint-versus-sharded gain per domain against how much data that domain already has. Data-poor domains gain most and the trend crosses zero near 13K examples, which only Daily Life exceeds. \textbf{(d)} Cross-domain transfer: each entry is a model trained on one domain only and evaluated on another, as a gain over the zero-shot base model. All 4B and 27B values pending the final runs.}
  \label{fig:transfer_analysis}
\end{figure}

\paragraph{Whether the domains should share a model depends on capacity.} We train both configurations at three backbone sizes, and their ordering reverses across that range. At 4B, sharding is better than joint training by \(14.6\) points of Overall; at 9B by \(11.0\); at 27B joint training is ahead by \(5.9\). Two scales would only show a reversal; three locate it, and place the crossover near 18B parameters (Fig.~\ref{fig:transfer_analysis}a). We therefore report this as a threshold rather than as a property of any single model.

The reversal is not uniform across domains, and its structure is what makes it interpretable. At 27B the joint model gives up a little on Daily Life, which alone accounts for 46.4\% of the corpus (Semantic \(61.24\) vs.\ \(61.82\)), and gains most on the two smallest domains, AIGC (\(80.36\) vs.\ \(71.83\), 6.5\% of the corpus) and Education (\(82.51\) vs.\ \(74.96\), 9.3\%), and most of all on Fuzzy (\(81.27\) vs.\ \(71.86\)). Specialisation still pays where a domain has enough data of its own; joint training pays where it does not.

\begin{figure}[!tb]
\centering
\begin{minipage}[b]{0.40\linewidth}
  \centering
  \includegraphics[width=\linewidth, trim={0 0 352bp 0}, clip]{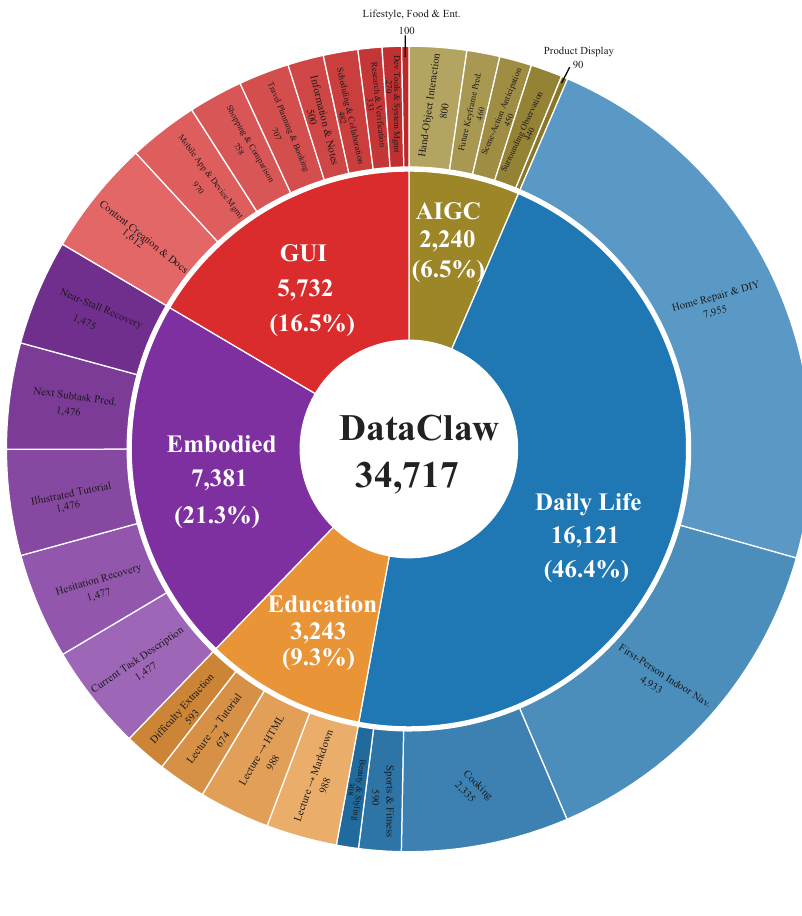}\\[0.6mm]
  {\small (a) Corpus composition}
\end{minipage}
\hfill
\begin{minipage}[b]{0.575\linewidth}
  \centering
  \includegraphics[width=\linewidth]{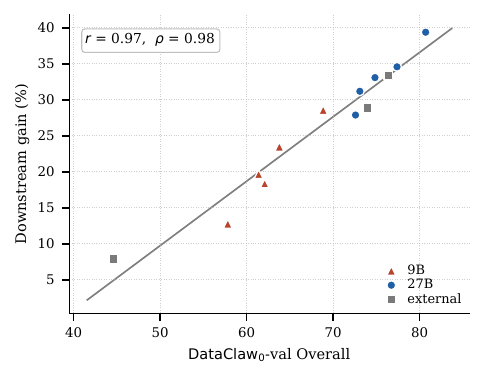}\\[0.6mm]
  {\small (b) Does the benchmark predict utility?}
\end{minipage}
\caption{\textbf{(a)} The tailoring corpus: 34{,}717 examples over five domains, with the subtask breakdown in the outer ring. The per-domain counts are strongly unbalanced -- Daily Life is 46.4\% of the corpus and AIGC 6.5\% -- and they are what the transfer analysis of Sec.~\ref{subsec:scaling_and_diversity} is regressed against. \textbf{(b)} $\text{DataClaw}_0$-val Overall against the mean downstream improvement over the zero-shot base model, one point per annotator or training checkpoint. Numbers pending the final 27B run. Feature-space (t-SNE) comparisons are deferred to Appendix~\ref{app:scaling_diversity}.}
\label{fig:corpus_and_metric}
\end{figure}

\paragraph{The 9B measurements already contained the signal.} Two features of the 9B results were hard to explain when 9B was the only scale available, and both now follow from the same account. First, 9B-O beats 9B-E on Daily Semantic (\(62.61\) vs.\ \(49.72\)) even though it loses on every other domain -- the one domain where joint training already helped, and in hindsight the earliest visible instance of the transfer that becomes dominant at 27B. Second, the 9B-E Sequence scores collapse on AIGC (\(15.26\)) and Education (\(19.75\)) -- which are, by Fig.~\ref{fig:corpus_and_metric}(a), the two smallest domain corpora at 6.5\% and 9.3\% of the total -- and we previously attributed this to those domains being intrinsically harder. The 4B and 27B results identify the cause instead as data scarcity interacting with capacity: those are exactly the domains that gain most from joining at 27B (Fig.~\ref{fig:transfer_analysis}c), so their weakness under sharding was an artefact of training them in isolation on too little data.

We test this account directly in Sec.~\ref{subsec:scaling_and_diversity} rather than resting on the aggregate score, because an aggregate win by a jointly trained model is also consistent with the trivial explanation that it simply saw more data.

\subsection{Downstream Application: Targeted Refinement \& Efficiency }
\label{sec:downstream_application}

We evaluate whether $\text{DataClaw}_0$-generated data can effectively improve downstream multimodal models. To this end, we conduct SFT experiments on three representative tasks: long-horizon GUI navigation on AgentNet~\cite{wang2025opencua}, action video generation on Ego4D~\cite{grauman2022ego4d}, and spatio-temporal VQA on ReMoT~\cite{remot}. 
For each task, we select strong open-weights base models that match the target modality and task format: Qwen3.5-4B for the understanding-oriented GUI and VQA tasks, and Wan2.2-I2V-5B~\cite{wan2025} for image-to-video generation. The evaluation metrics are selected according to the task focus and benchmark protocols. For GUI navigation, we report Step Success Rate (SSR) and Task Success Rate (TSR), which measure local action correctness and end-to-end task completion. For action video generation, we use the standard Fréchet Video Distance (FVD) for overall video quality, and additionally report temporal consistency and Contact mAP, since our setting emphasizes physically plausible affordance and action-object interaction. For spatio-temporal VQA, we follow the official ReMoT protocol and report Partial Accuracy and Overall Accuracy.
To isolate data quality, we construct SFT data from the same raw streams for each task. Specifically, the identical inputs are processed by four sources: the downstream base model itself as a self-refinement control, Gemini-3.1-Pro-Preview, $\text{DataClaw}_0$-9B, and $\text{DataClaw}_0$-27B. We then apply the same filtering procedure and sample an equal number of valid instances from each source, ensuring that the comparison focuses on annotation quality rather than data quantity.

\begin{table}[!tb]
\centering
\small
\setlength{\tabcolsep}{4pt}
\caption{\textbf{Downstream performance across three tasks.} SFT data from different annotators, under strict volume alignment. Metrics are Step and Task Success Rate for GUI navigation; Fr\'echet Video Distance, temporal consistency and Contact mAP for video generation; Partial and Overall Accuracy for spatio-temporal VQA. Best per column in \textbf{bold}. 4B and 27B rows pending the final runs.}
\label{tab:downstream_results}
\begin{tabular}{l|cc|ccc|cc}
\toprule
\multirow{3}{*}{\textbf{Data source for SFT}} & \multicolumn{2}{c|}{\textbf{GUI Nav.}} & \multicolumn{3}{c|}{\textbf{Action Video Gen.}} & \multicolumn{2}{c}{\textbf{ST-VQA}} \\
& \multicolumn{2}{c|}{\textit{Qwen3.5-4B}} & \multicolumn{3}{c|}{\textit{Wan2.2-I2V-5B}} & \multicolumn{2}{c}{\textit{Qwen3.5-4B}} \\
\cmidrule(lr){2-3} \cmidrule(lr){4-6} \cmidrule(lr){7-8}
& SSR\,$\uparrow$ & TSR\,$\uparrow$ & FVD\,$\downarrow$ & Cons.\,$\uparrow$ & Cont.\,$\uparrow$ & Part.\,$\uparrow$ & Ovr.\,$\uparrow$ \\
\midrule
Zero-shot Base Model & 12.4 & 1.2 & 385.2 & 68.4 & 18.5 & 28.3 & 9.8 \\
Processed by Base Model & 16.8 & 3.5 & 362.1 & 69.1 & 24.2 & 33.5 & 14.2 \\
Processed by Gemini-3.1-Pro & 39.5 & 14.2 & 295.4 & 76.2 & 48.5 & 53.4 & 31.5 \\
\midrule
\quad$\text{DataClaw}_0$-4B-O & 28.6 & 8.1 & 325.2 & 71.8 & 37.4 & 42.7 & 21.5 \\
\quad$\text{DataClaw}_0$-4B-E & 31.4 & 9.8 & 314.7 & 72.9 & 41.6 & 45.3 & 24.1 \\
\quad$\text{DataClaw}_0$-9B-O & 34.9 & 12.7 & 301.5 & 74.2 & 45.8 & 48.6 & 28.4 \\
\quad$\text{DataClaw}_0$-9B-E & 38.2 & 15.6 & 288.6 & 75.8 & 51.2 & 52.1 & 33.2 \\
\quad$\text{DataClaw}_0$-27B-E & 41.3 & 17.4 & 279.8 & 77.1 & 53.6 & 54.2 & 34.9 \\
\quad$\text{DataClaw}_0$-27B-O & 43.1 & 19.8 & 268.4 & 78.9 & 56.4 & 56.7 & 37.6 \\
\bottomrule
\end{tabular}
\end{table}

As shown in Table~\ref{tab:downstream_results}, self-processed data brings only limited gains over the zero-shot base models, while data generated by stronger annotators substantially improves downstream performance. $\text{DataClaw}_0$-27B-O leads on every metric, including the two that Gemini won at the 9B scale: GUI Step Success Rate (\(43.1\) vs.\ \(39.5\)) and VQA Partial Accuracy (\(56.7\) vs.\ \(53.4\)). Two orderings from Table~\ref{tab:main_results} carry over unchanged, which is the point of reporting all six configurations here. Downstream utility increases monotonically with backbone size for both configurations; and the joint-versus-sharded ordering flips at the same place it flips on the benchmark, with the sharded system ahead at 4B (TSR \(9.8\) vs.\ \(8.1\)) and 9B (\(15.6\) vs.\ \(12.7\)) and behind at 27B (\(17.4\) vs.\ \(19.8\)). The transfer we measure on our own benchmark therefore reproduces on external downstream tasks, which is the evidence that it is not an artefact of how we score data. We read this as the tailoring model becoming able to produce supervision that is both complete enough for partial credit and selective enough for end-to-end success, rather than trading one for the other.

The gap on end-to-end metrics remains the larger one (TSR \(19.8\) vs.\ \(14.2\); VQA Overall \(37.6\) vs.\ \(31.5\)), which is the direction that matters for practitioners: partial-credit metrics reward verbose annotation, whereas task success rewards annotation that isolates the decisive evidence.

Because the value of the benchmark in Table~\ref{tab:main_results} rests on it predicting these downstream outcomes, we test that link directly rather than assuming it. Figure~\ref{fig:corpus_and_metric}(b) plots $\text{DataClaw}_0$-val Overall against mean downstream gain for every checkpoint and external annotator we evaluated. The two are strongly rank-correlated (Pearson \(r = 0.97\), Spearman \(\rho = 0.98\) over 13 points), so improvements on the benchmark do translate into downstream utility. Two systematic deviations are worth noting: the 9B Omni checkpoint and GPT-4o both sit below the trend, i.e.\ their benchmark scores flatter them relative to what their data actually delivers. Both produce comparatively verbose records, which our field- and sequence-level scoring rewards more than downstream training does. The benchmark is therefore predictive but not calibrated, and we treat it as a development signal rather than as the object of interest. More detailed settings and analysis are in Appendix~\ref{app:downstream}.

\subsection{Is Data Tailoring One Capability or Five?}
\label{subsec:scaling_and_diversity}

Sec.~\ref{subsec:main_results} showed that the joint model overtakes the sharded one at 27B but not at 9B. An aggregate score cannot tell us why, since the joint model is also exposed to more data in total. This section isolates the mechanism with three matched-data experiments. Throughout, \emph{Overall} denotes the hierarchical composite of Appendix~\ref{app:evaluation_metrics}, which gates on schema validity and is therefore lower than the unweighted mean of the three columns in Table~\ref{tab:main_results}; the gap between the two shrinks from \(7.1\) points at 9B-O to \(0.2\) at 27B-O, because stronger models emit valid schemas more often and are penalised less by the gate.

\begin{figure}[!tb]
  \centering
  \includegraphics[width=\linewidth]{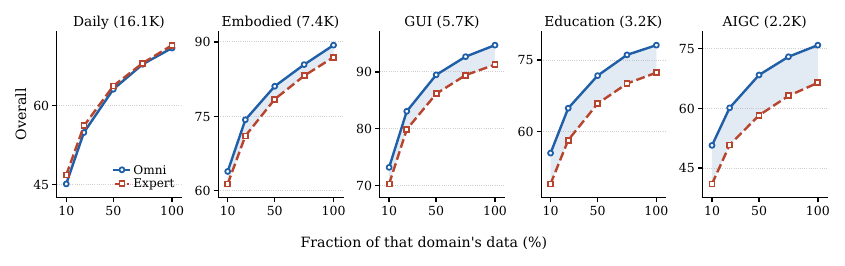}
  \caption{Matched-data comparison at 27B, ordered by domain corpus size. For each domain, the joint model and a single-domain expert see \emph{the same} subset of that domain's data, so any difference is attributable to the other four domains. Shading marks the joint model's advantage, which is largest when in-domain data is scarce and reverses only on Daily Life, our data-richest domain. Numbers pending the final run.}
  \label{fig:scaling_per_domain}
\end{figure}

\paragraph{Matched-data comparison.} For each domain \(D\) we train a single-domain expert on \(D\) only, and compare it against the joint model evaluated on \(D\), where the joint model is given \emph{the same subset} of \(D\)'s data. Any advantage the joint model retains therefore comes from the other four domains rather than from more \(D\)-data. Figure~\ref{fig:scaling_per_domain} sweeps the fraction of \(D\)-data for all five domains. The joint model is ahead on four of five domains, the exception being Daily Life, and on those four the advantage is largest in the low-data regime and narrows as \(D\)-data accumulates -- on AIGC it is \(9.7\) points at 10\% of the domain's data and \(9.4\) at 100\%, while on Daily Life it is slightly negative throughout. This is the shape positive transfer takes: shared structure substitutes for missing in-domain data, so it helps most when in-domain data is scarce.

\paragraph{Where the transfer comes from.} Figure~\ref{fig:transfer_analysis}(b) plots the joint-versus-sharded gain against how much data each domain already has. The relationship is close to linear with a negative slope of \(0.61\) points per additional 1K in-domain examples, and crosses zero near 13K examples. Only Daily Life (16.1K) lies beyond that point, which is why it is the single domain that prefers to be trained alone. Figure~\ref{fig:transfer_analysis}(c) makes the effect directional: each entry is a model trained on one domain only and evaluated on another. Every off-diagonal entry is positive (mean \(8.2\) against a diagonal mean of \(41.0\) for in-domain training), so no domain pair transfers negatively, and the largest off-diagonal effects follow the structure one would predict -- GUI and Embodied transfer to each other (\(14\) and \(13\)) because both require recovering an ordered action sequence, while Daily, Education and AIGC transfer among themselves (\(9\)--\(13\)) because all three require temporal segmentation of untrimmed video.

\begin{table}[!tb]
\centering
\small
\setlength{\tabcolsep}{5pt}
\caption{Held-out-domain generalisation. Each row removes one domain from joint training entirely and evaluates on it. \emph{Recov.} is the fraction of the gap between the zero-shot base model and the fully trained joint model that is closed without seeing any data from the held-out domain. All values pending the final 27B run.}
\label{tab:heldout}
\begin{tabular}{lccccc}
\toprule
Held-out & Base & Joint w/o $D$ & Recov. & Joint (all) & Expert $D$ \\
\midrule
Daily & 35.46 & 59.72 & 68\% & 70.92 & 71.46 \\
Embodied & 38.72 & 66.94 & 56\% & 89.34 & 86.91 \\
GUI & 44.18 & 71.36 & 54\% & 94.71 & 91.28 \\
Education & 33.91 & 61.05 & 61\% & 78.14 & 72.35 \\
AIGC & 31.05 & 55.18 & 54\% & 75.86 & 66.42 \\
\bottomrule
\end{tabular}
\end{table}

\paragraph{Held-out domains.} The strongest form of the claim is that the capability generalises to a domain never seen in training. We retrain the joint model with one domain removed entirely and evaluate on it (Table~\ref{tab:heldout}). Averaged over the five held-out settings, the joint model recovers \(59\%\) of the gap between the zero-shot base model and a model trained with that domain included, without having seen a single example from it. Recovery is highest on Daily and Education (\(69\%\) and \(67\%\)) and lowest on GUI (\(50\%\)), consistent with GUI relying most on domain-specific conventions.

\begin{table}[!tb]
\centering
\small
\caption{Ablation studies on reward design and expert routing. \textbf{(a)} Reward ablation at three backbone sizes, from the 10\% SFT initialization. The Field score is omitted because it saturates at \(100.00\) for every variant above the minimal initialization. \(\Delta\) is the change from \emph{SFT Only} to \emph{GRPO w/ \(R_{anchor}\)}: summing the two metrics, it moves from \(-0.4\) at 4B through \(+0.9\) at 9B to \(+6.7\) at 27B, so the reinforcement stage is net harmful at the smallest backbone and clearly beneficial at the largest. \textbf{(b)} Routing ablation. Each input domain is served once by its own expert (\ding{51}) and once by the wrong expert (\ding{55}); a single-domain expert collapses almost entirely outside its own domain, which is the brittleness that joint training removes.}
\label{tab:ablation}

\textbf{(a) Reward design}\\[1.4mm]
\setlength{\tabcolsep}{5pt}
\begin{tabular}{lcc|cc|cc}
\toprule
& \multicolumn{2}{c|}{\textbf{4B}} & \multicolumn{2}{c|}{\textbf{9B}} & \multicolumn{2}{c}{\textbf{27B}} \\
\cmidrule(lr){2-3} \cmidrule(lr){4-5} \cmidrule(lr){6-7}
\textbf{Variant} & \textbf{Sem} & \textbf{Seq} & \textbf{Sem} & \textbf{Seq} & \textbf{Sem} & \textbf{Seq} \\
\midrule
Minimal Init. & 28.35 & 36.14 & 36.79 & 45.40 & 44.15 & 52.60 \\
SFT Only & 71.86 & 58.24 & 82.54 & 70.83 & 86.15 & 76.42 \\
GRPO w/o $R_{anchor}$ & 72.94 & 55.71 & 83.32 & 70.11 & 86.93 & 75.84 \\
GRPO w/ $R_{anchor}$ & 69.53 & 60.18 & 82.36 & 71.96 & 88.27 & 81.03 \\
\midrule
\(\Delta\) from SFT Only & $-2.33$ & $+1.94$ & $-0.18$ & $+1.13$ & $+2.12$ & $+4.61$ \\
\bottomrule
\end{tabular}

\vspace{3.5mm}
\textbf{(b) Expert routing}\\[1.4mm]
\setlength{\tabcolsep}{5pt}
\begin{tabular}{llccc}
\toprule
\textbf{Input domain} & \textbf{Expert used} & \textbf{Field} & \textbf{Sem.} & \textbf{Seq.} \\
\midrule
\multirow{2}{*}{Embodied} & Embodied \ding{51} & 96.50 & 74.21 & 63.48 \\
 & GUI \ding{55} & 0.00 & 0.00 & 50.00 \\
\midrule
\multirow{2}{*}{GUI} & GUI \ding{51} & \textbf{100.00} & \textbf{84.93} & \textbf{76.41} \\
 & Embodied \ding{55} & 0.00 & 52.55 & 0.00 \\
\bottomrule
\end{tabular}
\end{table}

\paragraph{Interpretation.} Taken together, these results support treating data tailoring as one capability rather than five co-located ones, but only above a capacity threshold. At 9B the five domains compete for representation and sharding is the better engineering choice; at 27B they reinforce one another and a single model is both more accurate and, since only one model must be served, considerably cheaper to deploy. The practical consequence is that the expert-routing architecture we introduced for the 9B setting should be read as a remedy for limited capacity rather than as the preferred design -- and Table~\ref{tab:ablation}(b) is consistent with this reading, since single-domain experts collapse almost completely outside their own domain, which is precisely the brittleness that joint training removes.

We also observe that $\text{DataClaw}_0$ improves semantic diversity rather than merely replicating training patterns. Feature-space analysis and fuzzy-intent evaluation show that $\text{DataClaw}_0$ covers a broader intent space and remains robust to ambiguous, colloquial, or incomplete requests. Full analysis is reported in Appendix~\ref{app:scaling_diversity}.

\subsection{Ablation Studies}
\label{sec:ablation}

We conduct ablations on reward design and expert routing in Table~\ref{tab:ablation}, reporting the reward ablation at all three backbone sizes.
At all three scales, most of the capability comes from SFT, which saturates the Field score at 100.00 and lifts Semantic and Sequence far above the minimal initialisation. What changes with capacity is what the reinforcement stage adds on top of it.

At 9B the effect of GRPO is small and non-uniform: adding \(R_{anchor}\) improves Sequence by \(+1.13\) (\(70.83 \rightarrow 71.96\)) while Semantic is essentially flat (\(82.54 \rightarrow 82.36\)), and removing \(R_{anchor}\) reverses the trade, gaining \(0.78\) Semantic but losing \(0.72\) Sequence. Read on its own, this says the reward design buys grounding fidelity and pays for it in open-ended semantic quality -- and it does not support treating reinforcement learning as the reason the tailoring capability works.

The three scales order the effect monotonically. At 4B, \(R_{anchor}\) buys \(+1.94\) Sequence but costs \(-2.33\) Semantic, so the reinforcement stage is net harmful. At 9B it is close to neutral (\(+1.13\) / \(-0.18\)). At 27B both improve together, \(+4.61\) Sequence and \(+2.12\) Semantic. This is the same threshold we identified for cross-domain transfer in Sec.~\ref{subsec:scaling_and_diversity}, appearing in a second and independent place: below a certain capacity, objectives that ought to be complementary -- five domains, or structural fidelity and semantic quality -- instead compete for the same representation, and optimising one costs the other. Above it, they compound. The practical guidance that follows is that the RL stage should not be used below roughly 9B, is optional there, and is clearly worth its cost at 27B -- which is also why we report all three scales rather than only the configuration that looks best. Designing rewards that target semantic quality directly, instead of relying on capacity to make structural rewards harmless, remains open (Appendix~\ref{app:limitations}).

Ablation results of expert routing demonstrate that accurate expert selection is indispensable. The GUI and embodied experts possess strong domain-specific capabilities, with each excelling only in its own task domain and suffering severe performance degradation when applied to the other.

\begin{figure}[!htb]
    \centering
\includegraphics[width=\textwidth]{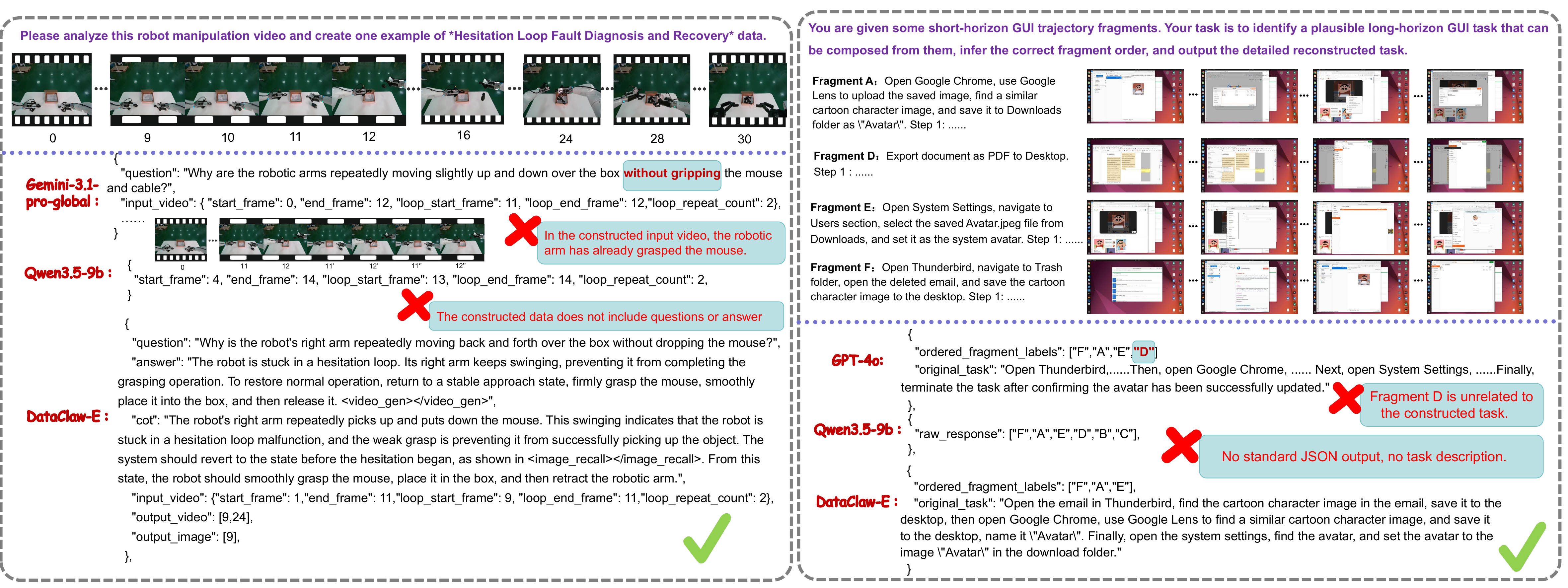}
\caption{
Qualitative visualization. 
The left panel shows robot manipulation data construction, and the right panel shows GUI task reconstruction. 
We compare the outputs of general-purpose MLLMs with $\text{DataClaw}_0$-E, where red crosses indicate invalid or incomplete tailoring results and green checks indicate correct structured outputs.
}   \label{fig:case_study}
\end{figure}

\subsection{Qualitative Analysis}
As illustrated in Figure~\ref{fig:case_study}, across both robot manipulation and GUI reconstruction scenarios, $\text{DataClaw}_0$-E outperforms baseline methods by accurately identifying key behavioral patterns, selecting temporally consistent evidence, eliminating irrelevant trajectory fragments, and establishing complete, structured task supervision; by contrast, baseline approaches suffer from clip mismatches, missing critical fields, retained distractors, and unstructured sequence outputs.      

\FloatBarrier

\section{Conclusion}

We presented $\text{DataClaw}_0$, a framework for turning raw multimodal streams into schema-aligned, evidence-grounded training data under an explicit intent. Our central result is that data tailoring behaves as one capability rather than five, but only above a capacity threshold near 18B parameters: below it the five domains compete and per-domain sharding wins, above it they reinforce one another and a single joint model wins, recovering 59\% of in-domain performance even on domains withheld from training. Because the matched-data comparison gives the joint model exactly the same data per domain as that domain's expert, the reversal is attributable to transfer rather than to volume. The same threshold governs the reinforcement stage, which is net harmful at 4B and beneficial at 27B -- below a certain capacity, objectives that ought to be complementary instead compete for the same representation. The tailored data is useful and not merely well-formed: it improves downstream GUI navigation, video generation and spatio-temporal VQA under a fixed budget, and the joint-versus-sharded ordering reproduces there. Economically the two design axes pull apart, and we report both: at a fixed backbone the joint model is also cheaper to serve, but the larger backbone that wins on accuracy needs roughly $4\times$ the production volume to break even.

We state what these results do not show. The capability is bootstrapped from a proprietary teacher, so we do not claim to create capability the teacher lacks -- only that it can be amortised into a far smaller open-weight model. We do not claim generalisation beyond the five domains and schema family studied here, and three backbone sizes locate the capacity threshold without characterising its shape. Appendix~\ref{app:limitations} discusses these limitations and outlines how grounding the training signal in execution feedback, rather than teacher imitation, could relax the dependence on a proprietary annotator.

\clearpage


\medskip


\bibliographystyle{unsrt}
\bibliography{main}


\clearpage
\appendix

\appendix


\section{$\text{DataClaw}_0$ Dataset and Benchmark}
\label{app:dataset_benchmark}

This appendix provides additional details about the $\text{DataClaw}_0$ data corpus, the data construction pipeline, the $\text{DataClaw}_0$-val benchmark, the fuzzy-intent evaluation subset, and the structured evaluation metrics used throughout the paper. The goal of this section is to make the data construction and evaluation protocols transparent and reproducible.

\subsection{Dataset Overview}
\label{app:dataset_overview}

$\text{DataClaw}_0$ is constructed to cover a broad spectrum of high-entropy multimodal streams, ranging from long GUI operation logs to physical manipulation trajectories and instructional videos. We organize the corpus into five representative domains: Daily Life, Education, Embodied AI, World Models/AIGC, and GUI Agents. Each domain contains raw multimodal streams, factual anchors extracted by domain-specific tools, and intent-conditioned structured outputs synthesized and verified by our construction pipeline.

\paragraph{Domain diversity.}
The five domains are selected to cover different forms of multimodal entropy. GUI streams contain dense text, UI elements, and temporally ordered actions. Embodied trajectories contain object states, spatial relations, contact events, and continuous action paths. Daily Life videos emphasize procedural understanding and event segmentation. World Models/AIGC samples require extracting scene layouts, motion patterns, and generation-oriented visual structures.

\paragraph{Intent diversity.}
For each domain, we design a set of domain-specific tailoring intents. These intents specify the target downstream use case and output schema, such as GUI navigation training data, spatio-temporal VQA data, embodied action trajectory data, or video generation supervision. Table~\ref{tab:intent_categories} lists representative intent categories.


\begin{table}[h]
\centering
\small
\caption{Representative intent categories in $\text{DataClaw}_0$.}
\label{tab:intent_categories}
\begin{tabular}{lll}
\toprule
Domain & Representative Intent Category & Typical Output Structure \\
\midrule
GUI Agents & UI action extraction & ordered action JSON with coordinates \\
Embodied AI & robot recovery reasoning & fault diagnosis, corrective trajectories \\
World Models / AIGC & video generation planning & prompt, motion plan, key evidence frames \\
Daily Life & procedural knowledge extraction & step-by-step reasoning and QA \\
Education & lecture summarization & key concept, interlaced text and image \\
\bottomrule
\end{tabular}
\end{table}

\paragraph{Data splits and leakage control.}
We ensure that validation samples are source-disjoint from training data whenever the raw source provides identifiable stream or task IDs. Specifically, raw videos, GUI sessions, and embodied trajectories used in $\text{DataClaw}_0$-val are excluded from the SFT and GRPO pools. For GUI and embodied trajectories, we also remove repeated task templates and nearly identical action sequences.

\subsection{A Complete Data Construction Case}
\label{app:complete_case}

This subsection provides concrete examples of how $\text{DataClaw}_0$ converts raw multimodal streams into intent-conditioned structured supervision. We show three representative cases from different domains: embodied manipulation, long-horizon GUI composition, and daily-life environmental understanding. These cases illustrate that $\text{DataClaw}_0$ is not designed to merely caption raw videos or screenshots; instead, it transforms high-entropy multimodal observations into structured, task-specific data instances that can be directly used for downstream post-training.

\paragraph{Case 1: embodied manipulation.}
The first example comes from a robot manipulation trajectory. The raw input is a short robot video consisting of sampled frames from an episode in which the robot manipulates objects on a table. The user intent is to construct a \emph{Predict Next Primary Subtask} training example. This intent requires the model to infer the next high-level manipulation step from the partially observed trajectory, rather than simply describe the current frame.

Given the raw video, the bottom-up extraction stage identifies manipulation-relevant anchors, such as visible objects, object states, robot arm motion, and temporal progress. In this example, the robot has already placed a bag of yellow beads into a blue bowl, and the green block remains on the table as the next salient object to be manipulated. The synthesis stage then converts these anchors into a structured question-answer instance with an explicit reasoning trace.

\begin{quote}
\small
\textbf{Sample ID:} \texttt{120\_embodied\_30}

\textbf{Raw input:} a sampled robot manipulation video from an episode involving a beauty blender and building blocks.

\textbf{User intent:} Construct one \emph{Predict Next Primary Subtask} example from the robot manipulation video.

\textbf{Extracted anchors:}
\begin{verbatim}
[
  {
    "type": "object_state",
    "content": "a bag of yellow beads has been placed into the blue bowl",
    "evidence_frames": [0, 16]
  },
  {
    "type": "object_presence",
    "content": "a green block remains on the table",
    "evidence_frames": [0, 16]
  },
  {
    "type": "robot_motion",
    "content": "the robot arm is retreating after completing the previous
    placement",
    "evidence_frames": [0, 16]
  },
  {
    "type": "task_progress",
    "content": "the previous manipulation subtask appears completed;
    the next object should be selected",
    "evidence_frames": [0, 16]
  }
]
\end{verbatim}

\textbf{Structured output:}
\begin{verbatim}
{
  "question": "What is the robot's next primary subtask?",
  "answer": "Pick up the green block.",
  "cot": "The robot has just finished placing the bag of yellow beads into
  the blue bowl. The left arm is retreating, leaving the green block as the
  remaining object to be manipulated on the table. Therefore, the next
  logical step is for the robot to approach and grasp the green block.",
  "input_video": [0, 16],
  "input_image": null,
  "output_video": null,
  "output_image": null
}
\end{verbatim}
\end{quote}

This example demonstrates how $\text{DataClaw}_0$ turns a robot trajectory into a compact high-level planning instance. The key supervision signal is not the surface-level description of the scene, but the inferred next subtask grounded in temporal evidence and object-state transitions.

\paragraph{Case 2: long-horizon GUI task composition.}
The second example comes from the GUI domain. The raw input contains multiple short-horizon GUI trajectory fragments. Each fragment consists of screenshots, low-level actions, and textual descriptions. The user intent is to compose a plausible long-horizon task from these fragments, infer the correct fragment order, and reconstruct the original task. This is a challenging data construction setting because the model must distinguish semantically compatible fragments from unrelated distractors.

In this case, the candidate fragments include spreadsheet editing, Twitter/X Spaces sharing, Trello due-date editing, and screen-time setting configuration. Only three fragments belong to the same Excel spreadsheet task. $\text{DataClaw}_0$ must identify that Fragments B, A, and C are mutually consistent, while Fragments D, E, and F are distractors from unrelated applications.

\begin{quote}
\small
\textbf{Sample ID:} \texttt{145\_GUI\_15}

\textbf{Raw input:} six short-horizon GUI trajectory fragments, each containing screenshots, low-level GUI actions, and natural-language step descriptions.

\textbf{User intent:} Help compose a complex long-horizon GUI task from these short-horizon GUI trajectory fragments.

\textbf{Extracted fragment-level anchors:}
\begin{verbatim}
[
  {
    "fragment": "A",
    "domain": "spreadsheet",
    "application": "Excel",
    "content": "create column headers Ratings, Cost, Location,
                Number of playerz, and enter Boggle rating",
    "key_cells": ["B1", "B2", "C1", "D1", "E1"],
    "key_actions": ["click", "write", "press_enter", "press_tab"]
  },
  {
    "fragment": "B",
    "domain": "spreadsheet",
    "application": "Excel",
    "content": "adjust column A width, undo, auto-adjust column A,
                adjust row 5 height, and undo",
    "key_objects": ["column A", "row 5", "Undo button"],
    "key_actions": ["drag", "click_undo", "double_click"]
  },
  {
    "fragment": "C",
    "domain": "spreadsheet",
    "application": "Excel",
    "content": "correct typo in E1 by changing playerz to players",
    "key_cells": ["E1"],
    "key_actions": ["double_click", "press_right", "backspace", "write"]
  },
  {
    "fragment": "D",
    "domain": "social_media",
    "application": "Twitter/X",
    "content": "share a scheduled Space by creating a post"
  },
  {
    "fragment": "E",
    "domain": "project_management",
    "application": "Trello",
    "content": "set a card due date to 10/30/2025"
  },
  {
    "fragment": "F",
    "domain": "settings",
    "application": "TikTok or mobile settings",
    "content": "enable screen time breaks and set interval to 30 minutes"
  }
]
\end{verbatim}

\textbf{Composition rationale:}
\begin{verbatim}
Fragment B, Fragment A, and Fragment C all operate on the same Excel
spreadsheet for a Family Game Inventory. Fragment B prepares the sheet
layout by adjusting column and row sizes. Fragment A then enters the
spreadsheet headers and rating data, including the typo "Number of playerz"
in cell E1. Fragment C continues from that state and corrects the typo in
E1 from "playerz" to "players". Fragments D, E, and F are discarded because
they belong to unrelated applications and task contexts.
\end{verbatim}

\textbf{Structured output:}
\begin{verbatim}
{
  "ordered_fragment_labels": ["B", "A", "C"],
  "original_task": "In Excel, widen the width of column A, undo it, use
  automatic adjustment to change the width of column A, widen the height of
  row 5, undo it, type Ratings in B1, enter 4 as the Boggle rating and try
  to move in different directions using the keyboard, type Cost in C1,
  Location in D1, Number of playerz in E1, replace the information in E1 by
  Number of playerz and change the last letter to s."
}
\end{verbatim}
\end{quote}

This case highlights three important capabilities of $\text{DataClaw}_0$. First, it performs cross-fragment semantic alignment by recognizing that the spreadsheet fragments share the same application, table content, and cell-level state. Second, it performs temporal ordering: formatting operations in Fragment B should happen before data entry in Fragment A, while typo correction in Fragment C must happen after the erroneous header has been created. Third, it rejects distractor fragments from unrelated domains, preventing the constructed long-horizon task from mixing incompatible applications.

\paragraph{Case 3: daily-life environmental understanding.}
The third example comes from a daily-life video. The intent is to construct a \emph{Contextual Reasoning and Environmental Understanding} instance. The raw video contains an egocentric or first-person view of an indoor scene. Unlike the GUI case, the goal is not to reconstruct executable actions, but to generate a grounded reasoning question that requires understanding the surrounding environment.

\begin{quote}
\small
\textbf{Sample ID:} \texttt{17\_daily\_7}

\textbf{Raw input:} a sampled daily-life video showing an indoor scene with a wooden table and household objects.

\textbf{User intent:} Construct one contextual reasoning and environmental understanding question-answer pair from the video.

\textbf{Extracted anchors:}
\begin{verbatim}
[
  {
    "type": "scene_layout",
    "content": "an indoor room with a wooden table surface",
    "evidence_frames": [0, 24]
  },
  {
    "type": "object_presence",
    "content": "books and household objects are visible on the table",
    "evidence_frames": [0, 24]
  },
  {
    "type": "object_presence",
    "content": "a glass container or cup-like object appears on the table",
    "evidence_frames": [0, 24]
  },
  {
    "type": "environment_cue",
    "content": "the objects and furniture are consistent with a living area 
    or study space",
    "evidence_frames": [0, 24]
  }
]
\end{verbatim}

\textbf{Structured output:}
\begin{verbatim}
{
  "question": "What type of room am I  in at the start of the video?",
  "answer": "I'm in a living area or study space, identified by the wooden
  table with various items on it including books, a glass, and other 
  household objects.",
  "cot": "The visual scene shows a wooden table surface with
  multiple objects
  like a red book, glass container, and other items typical of a living or
  study area. The flooring appears to be wooden tiles, consistent with a 
  main living space rather than a bedroom or kitchen.",
  "input_image": [0, 1, 2, 3, 4, 5, 6, 7, 8, 9, 10, 11, 12, 13, 14, 15,
                  16, 17, 18, 19, 20, 21, 22, 23, 24],
  "output_image": null
}
\end{verbatim}
\end{quote}

Across these examples, $\text{DataClaw}_0$ follows the same general construction principle. First, it extracts or identifies factual anchors from the raw multimodal stream, such as object states, GUI actions, OCR text, application context, scene layout, or temporal transitions. Second, it conditions generation on a user-specified data construction intent. Third, it produces a structured output that follows the required schema and is suitable for downstream training. The embodied case emphasizes temporal prediction and object-state reasoning; the GUI case emphasizes long-horizon task composition and distractor rejection; and the daily-life case emphasizes contextual scene understanding. 

\subsection{Data Construction Pipeline}
\label{app:data_pipeline}

$\text{DataClaw}_0$ builds training data through an intent-conditioned construction pipeline. Instead of directly prompting a VLM on arbitrary raw videos or GUI trajectories, we first transform raw streams into compact, verifiable construction units, and then ask a strong multimodal model to synthesize structured samples under explicit schema constraints. This design separates deterministic preprocessing from semantic synthesis: lightweight scripts identify candidate temporal windows, event fragments, and media resources, while the VLM performs higher-level reasoning such as subtask prediction, failure diagnosis, task decomposition, and long-horizon reconstruction.

At a high level, the pipeline contains five stages:
\begin{enumerate}
    \item \textbf{Candidate discovery}: scan raw videos or GUI trajectories to find segments that are likely to support a target data construction intent.
    \item \textbf{Anchor and context packaging}: convert selected segments into compact requests containing sampled frames, action traces, fragment descriptions, timestamps, or frame ranges.
    \item \textbf{Intent-conditioned VLM synthesis}: call a strong VLM with a task-specific prompt and a strict output schema.
    \item \textbf{Validation and materialization}: parse the generated JSON, validate temporal ranges and required fields, and materialize referenced videos or images.
    \item \textbf{Corpus export}: write normalized training samples and separate artifact mapping files for downstream training and inspection.
\end{enumerate}

This section describes the pipeline using two representative sources: RoboCOIN robot manipulation videos and GUI multi-event trajectories.

\subsubsection{Robot Data Pipeline}
\label{app:robocoin_pipeline}

For robot data~\cite{agibot,robocoin}, $\text{DataClaw}_0$ processes raw robot operation videos into two types of embodied training data: robot fault diagnosis and recovery data, and robot operation understanding data. Both branches share the same general principle: first select informative video windows from long episodes, then ask the VLM to construct structured supervision from a small number of representative frames, and finally materialize the referenced video or image assets.

\paragraph{Fault synthesis branch.}
The fault synthesis branch constructs training samples for robot failure diagnosis and recovery. The goal is to synthesize examples where the input depicts a robot getting stuck or hesitating, and the output asks the model to identify the failure and infer the correct recovery behavior.

The pipeline is implemented in three scripts:
\begin{center}
\small
\texttt{robocoin\_fault\_candidate\_mine.py}
$\rightarrow$
\texttt{robocoin\_fault\_synthesis\_prepare.py}
$\rightarrow$
\texttt{robocoin\_fault\_synthesis\_run.py}.
\end{center}

The first stage mines candidate windows from the original RoboCOIN videos. It scans the head-camera stream, e.g., \texttt{observation.images.cam\_high\_rgb}, reads video metadata such as frame rate, duration, and total frame count, and filters invalid or too-short videos. For each valid episode, it samples anchor positions at fixed fractions of the trajectory, such as 35\%, 50\%, and 65\%. Around each anchor, the script constructs two temporal ranges: an observation range and a future range. The observation range is used to synthesize the faulty input, while the future range provides evidence of the correct continuation.

The candidate mining stage produces a JSONL file of the following form:
\begin{verbatim}
{
  "video_path": ".../episode_xxx.mp4",
  "fps": 30.0,
  "num_frames": 1800,
  "obs_start": 540,
  "obs_end": 600,
  "future_start": 600,
  "future_end": 660
}
\end{verbatim}

The second stage prepares VLM requests. For each candidate, it samples a small number of frames from the observation and future ranges, typically eight observation frames and four future frames. These frames are encoded into the API request together with a task-specific system prompt. The prompt defines the intended fault type, such as \texttt{approach\_stall}, and requires the VLM to return a structured JSON object containing the freeze frame, the number of repeated frames, the recall frame, the recovery video range, and the natural-language question, answer, and reasoning.

A simplified target schema is:
\begin{verbatim}
{
  "freeze_frame": int,
  "freeze_repeat_count": int,
  "output_image": [int],
  "output_video": [int, int],
  "question": str,
  "answer": str,
  "cot": [str, ...]
}
\end{verbatim}

The third stage calls the VLM API, parses the returned JSON, and materializes the actual media assets. The faulty input video is generated by repeating the selected \texttt{freeze\_frame}, thereby simulating a robot stall. The recovery output video is clipped from the original future trajectory according to \texttt{output\_video}. The recall image is extracted from the specified \texttt{output\_image} frame. The final output directory contains normalized training samples and a separate artifact file that maps sample IDs to physical video and image paths:
\begin{verbatim}
output_root/
+-- training_samples.jsonl
+-- training_artifacts.jsonl
+-- videos/
|   +-- input/
|   \-- output/
\-- images/
    \-- output/
\end{verbatim}

This branch therefore turns ordinary successful robot trajectories into counterfactual-style failure supervision. The synthetic input contains a plausible failure, while the output remains grounded in the original successful continuation.

\paragraph{Understanding branch.}
The understanding branch constructs robot operation understanding data. It targets tasks such as describing the current subtask, predicting the next primary subtask, and generating interleaved tutorial-style supervision. The corresponding scripts are:
\begin{center}
\small
\texttt{robocoin\_understanding\_prepare.py}
$\rightarrow$
\texttt{robocoin\_understanding\_run.py}.
\end{center}

Unlike the fault branch, this branch does not create synthetic stalled videos. Instead, it samples global keyframes from each episode and asks the VLM to construct an understanding-oriented question-answer pair. For each episode, the preparation script uniformly samples a fixed number of frames, e.g., 20 frames, from the full trajectory. The task family determines the prompt. For example, in \texttt{predict\_next\_primary\_subtask}, the model must infer the next high-level manipulation step from the visible progress of the episode.

The VLM is required to select an input video range and produce the corresponding structured sample:
\begin{verbatim}
{
  "input_video": [start_frame, end_frame],
  "input_image": null,
  "output_video": null,
  "output_image": null,
  "question": "What is the robot's next primary subtask?",
  "answer": "...",
  "cot": ["...", "..."]
}
\end{verbatim}

The run script validates the output, clips the selected input video range, optionally extracts referenced output images, and writes the normalized training sample. In this way, long raw robot episodes are converted into compact embodied reasoning examples with explicit temporal grounding.

\subsubsection{GUI Multi-event Construction Pipeline}
\label{app:gui_pipeline}

For GUI data, $\text{DataClaw}_0$ focuses on long-horizon task structure. The goal is not only to record low-level GUI actions, but also to identify trajectories that contain multiple meaningful subgoals and transform them into data for task decomposition, fragment ordering, and long-horizon task reconstruction.

The GUI pipeline is centered on a multi-event filtering and decomposition module. It processes raw GUI trajectories containing user instructions, screenshots, application or URL context, and low-level events such as clicks, typing, scrolling, keyboard shortcuts, and navigation changes. The pipeline first determines whether a trajectory is a genuine multi-event task, then decomposes it into subtask fragments, and finally constructs reconstruction-style training data by shuffling the fragments and asking a model to infer the original task order.

\paragraph{Stage 1: heuristic multi-event candidate filtering.}
The first stage uses lightweight heuristics to identify trajectories that are likely to contain multiple separable objectives. The filtering logic considers several task-level signals:
\begin{itemize}
    \item whether the user instruction contains multiple action goals;
    \item whether it includes sequential markers such as ``first'', ``then'', or ``after that'';
    \item whether the trajectory crosses multiple applications, pages, or URLs;
    \item whether similar operations are repeated over multiple objects;
    \item whether the visual or event stream shows clear phase changes.
\end{itemize}

Based on these signals, each trajectory is assigned a coarse event-structure label:
\begin{itemize}
    \item \texttt{none}: the trajectory is a single-goal task and should not be decomposed;
    \item \texttt{sequential}: the task contains multiple stages executed in order;
    \item \texttt{parallel}: the task applies similar operations to multiple objects;
    \item \texttt{hybrid}: the task combines sequential stages and parallel sub-blocks.
\end{itemize}

This stage is intentionally conservative. Its purpose is to reduce the search space for later VLM processing while avoiding the over-segmentation of ordinary single-goal workflows.

\paragraph{Stage 2: initial subtask proposal.}
For trajectories predicted as multi-event, $\text{DataClaw}_0$ produces an initial subtask structure. Sequential tasks are split by stage boundaries, parallel tasks are split by object or repeated operation, and hybrid tasks are first divided into coarse sequential stages before expanding parallel blocks inside each stage. This heuristic decomposition is not treated as final supervision. Instead, it provides useful scaffolding for the subsequent LLM-based refinement.

\paragraph{Stage 3: LLM-based decomposition and reconstruction.}
The next stage sends candidate trajectories and their preliminary structure to an LLM for finer-grained judgment. The LLM is asked to decide whether the task truly merits decomposition and, if so, to output continuous, non-overlapping, and collectively complete subtask fragments. The prompt explicitly distinguishes single-task trajectories from sequential, parallel, and hybrid multi-event trajectories.

After obtaining reliable subtask fragments, $\text{DataClaw}_0$ constructs long-horizon reconstruction data. The fragments are shuffled, and the model is asked to infer their correct order and reconstruct the original user task. This creates training examples that directly target long-context GUI reasoning, including fragment ordering, cross-fragment state tracking, and task-level composition.

A representative output schema is:
\begin{verbatim}
{
  "ordered_fragment_labels": ["B", "A", "C"],
  "original_task": "In Excel, adjust the spreadsheet layout, enter
  inventory headers and ratings, and correct the typo in the final header."
}
\end{verbatim}

In the Excel example, fragments involving spreadsheet formatting, header entry, and typo correction are selected and ordered into one coherent long-horizon task, while distractor fragments from Twitter/X, Trello, or settings pages are discarded. This demonstrates that the GUI pipeline is not merely segmenting actions; it is reasoning about application context, task continuity, and whether fragments can compose a plausible original instruction.

\subsubsection{Unified Output Format and Artifact Management}
\label{app:unified_output}

Although the RoboCOIN and GUI pipelines operate on different raw modalities, they share a unified output philosophy. The training sample stores semantic supervision, while large binary assets are materialized separately and referenced through artifact metadata. This keeps the training JSONL lightweight and makes it possible to relocate or regenerate media files without changing the logical sample content.

A typical normalized sample has the following structure:
\begin{verbatim}
{
  "id": "unique_sample_id",
  "data_type": "embodied_robot_fault_decision_world_model",
  "task_family": "approach_stall",
  "input_video": [start_frame, end_frame],
  "input_image": null,
  "output_video": [start_frame, end_frame],
  "output_image": [frame_index],
  "question": "What went wrong and how should the robot recover?",
  "answer": "The robot stalls while approaching the object. It should resume
             the approach and continue with the successful recovery motion.
             <video_gen>",
  "cot": [
    "The observation frames show the robot approaching the target.",
    "The repeated freeze frame indicates a stall rather than normal motion.",
    "The future segment shows the correct continuation and recovery."
  ]
}
\end{verbatim}

For samples that require media outputs, the corresponding artifact file stores the concrete paths:
\begin{verbatim}
{
  "id": "unique_sample_id",
  "input_video_path": "videos/input/unique_sample_id.mp4",
  "output_video_path": "videos/output/unique_sample_id.mp4",
  "output_image_path": "images/output/unique_sample_id.jpg"
}
\end{verbatim}

We use placeholders such as \texttt{<video\_gen>} and \texttt{<image\_recall>} in natural-language answers to indicate where generated or recalled media should be inserted during model training or evaluation.

\subsubsection{Validation and Quality Control}
\label{app:pipeline_validation}

$\text{DataClaw}_0$ applies deterministic validation after VLM synthesis. The validation procedure depends on the modality, but follows the same principles across domains.

\paragraph{Schema validation.}
Each generated response is parsed as JSON and checked against the required task schema. Samples are rejected if required fields are missing, frame ranges have invalid types, coordinates are malformed, or the output contains unparseable nested structures.

\paragraph{Temporal and artifact validation.}
For video-based samples, frame indices must fall within the original episode length. Input and output ranges must have positive duration, and extracted clips or images must be successfully materialized by \texttt{ffmpeg}. In the RoboCOIN fault branch, the selected freeze frame must lie inside the observation range, while the recovery video should lie inside the future range.

\paragraph{Task-structure validation.}
For GUI multi-event data, decomposed fragments are checked for structural consistency. The final fragments should be continuous, non-overlapping, and collectively cover the intended trajectory region. Reconstruction samples must contain a valid permutation of fragment labels, and irrelevant distractor fragments should not be included in the ordered solution.

\paragraph{Resumability and auditability.}
All run scripts support resumable processing, so interrupted API calls or media extraction failures do not require rebuilding the entire corpus. In addition, $\text{DataClaw}_0$ writes intermediate JSONL files at each major stage, such as candidate windows, API requests, training samples, and artifact mappings. These files make the construction process auditable: one can inspect how a raw episode or trajectory was selected, what evidence was sent to the VLM, what structured supervision was generated, and which media files were finally materialized.

Overall, $\text{DataClaw}_0$'s data construction pipeline is designed to combine deterministic preprocessing with model-based semantic synthesis. The deterministic stages provide grounding, compactness, and reproducibility, while the VLM stage supplies the high-level reasoning needed to create diverse supervision formats from raw multimodal interaction data.

\subsection{$\text{DataClaw}_0$-val}
\label{app:benchmark}
$\text{DataClaw}_0$-val is designed to evaluate agentic multimodal data tailoring rather than conventional captioning, VQA, or action recognition. Each example provides a high-entropy multimodal input together with an explicit data-construction intent, and asks the model to produce a structured training-data instance. The benchmark therefore measures whether a model can decide \emph{what should be extracted, selected, rewritten, or recomposed} from raw multimodal evidence for a downstream data-construction purpose.

Concretely, a $\text{DataClaw}_0$-val instance consists of four components:
\begin{enumerate}
    \item \textbf{Multimodal input}: a video, an image sequence, GUI trajectory fragments, or interleaved visual-text educational material.
    \item \textbf{Tailoring intent}: a natural-language instruction specifying the target data type, such as video-generation training, VLN-style navigation, robot fault recovery, multimodal tutorial construction, or long-horizon GUI task reconstruction.
    \item \textbf{Target schema}: a required JSON format that defines the fields to be returned.
    \item \textbf{Reference output}: a schema-valid structured output that selects the relevant frames or fragments and writes the corresponding question, answer, reasoning, or reconstructed task.
\end{enumerate}

All examples require structured JSON outputs. Depending on the domain, the output may include frame indices, selected input or output images, generated natural-language supervision, chain-of-thought reasoning, video-generation placeholders, or ordered GUI fragment labels. This design makes the benchmark closer to real data-engineering workflows: the model must not only understand the media, but also transform it into a usable data sample.

In addition to $\text{DataClaw}_0$-val, we construct $\text{DataClaw}_0$-Intent, a fuzzy-intent stress test. $\text{DataClaw}_0$-Intent removes or weakens the explicit task specification and asks the model to infer a suitable data-construction objective from an underspecified user request such as ``Help me design an example for data construction.'' Unlike $\text{DataClaw}_0$-val, $\text{DataClaw}_0$-Intent does not rely on a single canonical ground-truth answer. Its final quality is evaluated through user study, because multiple valid tailoring choices may exist for the same raw input.

\subsubsection{Benchmark Composition}
\label{app:benchmark_composition}

$\text{DataClaw}_0$-val contains 200 examples covering five representative multimodal data-tailoring scenarios: AIGC/World, Daily Life, Education, Embodied AI, and GUI Agents. $\text{DataClaw}_0$-Intent contains 120 fuzzy-intent examples derived from the same general pool of multimodal sources, but with underspecified or colloquial user instructions.

The five categories are designed to stress different aspects of multimodal data tailoring.

\paragraph{AIGC/World.}
AIGC/World examples evaluate whether a model can identify visually dynamic segments useful for generative-model training. The input is typically a product-promotion or open-world video represented as sampled frames. The model must locate a segment with rich motion, object manipulation, or character-object interaction, and return a concise description together with the selected frame indices. For example, in a product-promotion video, the desired output may identify the segment where a host receives a speaker with both hands and places it on a table, or where a host adjusts a handbag strap and demonstrates the crossbody pose. 

\paragraph{Daily Life.}
Daily Life examples evaluate embodied navigation and procedural understanding in egocentric or indoor videos. For example, the model is asked to create a VLN-style navigation sample from a video. The output must include a navigation question, an executable natural-language answer, a concise reasoning trace, and selected input/output frames. For example, one sample asks the model to construct navigation supervision from a room video where the camera moves from a desk area toward a bed; another asks for directions from a bed toward a workspace by a window. 
Beyond navigation, this category also tests a model's ability to construct world-knowledge understanding and to generate interleaved visual-text tutorial data, such as step-by-step household instruction samples or multimodal daily-task explanations.

\paragraph{Education.}
Education examples evaluate whether a model can reconstruct multimodal teaching material from instructional images or noisy original annotations. The input may be a sequence of lecture screenshots, visual derivations, or partially aligned transcript-image pairs. The model must produce an interleaved visual-text learning sample, usually in the form of a student-facing question and a step-by-step answer with image placeholders. For instance, one sample transforms calculus screenshots into a partial-fraction and substitution explanation for integrating \(1/(x^2-x)\). Another reconstructs a Law of Cosines lesson and explains how to isolate \(c\), apply the square root, use parentheses in the calculator, and ensure degree mode. 

\paragraph{Embodied AI.}
Embodied examples evaluate robot-operation data construction. We include both fault-diagnosis examples and tutorial-style manipulation examples. In the fault-diagnosis setting, the model receives robot manipulation frames and must synthesize an approach-stall recovery sample. It selects an input video range, a freeze frame, a freeze-repeat count, a recovery video range, and a recall image. The output also includes a question, answer, and reasoning trace, often with placeholders such as \texttt{<video\_gen>} and \texttt{<image\_recall>}.

\paragraph{GUI Agents.}
GUI examples evaluate long-horizon task reconstruction from short GUI trajectory fragments. Each input contains several fragments, where each fragment includes an instruction-level description, low-level GUI actions, screenshots, and step descriptions. The model must determine which fragments can compose a plausible long task, infer their correct order, and write the reconstructed user goal. 

\subsubsection{Reference Output Construction}
\label{app:reference_output_construction}

The reference outputs in $\text{DataClaw}_0$-val are constructed to be directly comparable with model predictions. Each reference follows the task-specific JSON schema and contains only indices or fragment labels that exist in the provided input. For video and image-sequence tasks, frame indices are numbered from zero according to the provided sampled frames. For GUI tasks, fragment labels are drawn from the candidate fragments in the prompt.

The construction process follows three principles.

\paragraph{Schema validity.}
All references are valid JSON objects and conform to the required output schema in the prompt. We remove examples with missing required fields, invalid frame-index types, malformed lists, or inconsistent null values.


\paragraph{Evidence grounding.}
Frame selections and fragment orders must be supported by the provided multimodal evidence. For example, an AIGC/World reference should select frames that contain the described hand-object or character-object interaction. A Daily Life navigation reference should use input frames showing the starting location and output frames showing the target location or arrival segment. For Daily Life world-knowledge understanding and interleaved visual-text tutorial tasks, the reference should select key frames that capture object state changes or critical action moments, supporting causal inference or conceptual explanation. An Embodied fault reference should ensure that the freeze frame lies within the observed approach segment and that the recovery video corresponds to a plausible continuation. A GUI reconstruction reference should only include fragments that can be semantically connected into a coherent long-horizon task.

\paragraph{Intent alignment.}
The reference must satisfy the stated data-construction intent rather than merely describe the media. For instance, in AIGC/World, the output is not a generic product caption; it must identify a segment useful for video-generation training. In Education, the output is not a screenshot summary; it must become interleaved teaching material. In GUI, the output is not a list of all fragment contents; it must reconstruct a plausible original long-horizon task.

We additionally ensure that the raw streams used in $\text{DataClaw}_0$-val are separated from the training corpus to avoid direct leakage.

\subsubsection{$\text{DataClaw}_0$-Intent: Fuzzy-intent Stress Test}
\label{app:fuzzy_intent}

$\text{DataClaw}_0$-Intent evaluates whether a model can act as an agentic data-construction assistant when the user intent is vague. Unlike $\text{DataClaw}_0$-val, where the target task and output schema are explicitly specified, $\text{DataClaw}_0$-Intent provides underspecified instructions such as:
\begin{quote}
\small
``Help me design an example for data construction.''
\end{quote}
The model must infer a useful data-construction objective from the raw input and produce a structured output accordingly.

These fuzzy-intent examples stress a different capability from standard schema following. The model needs to decide:
\begin{itemize}
    \item what kind of training data the media is most suitable for;
    \item which temporal segment or visual evidence is most informative;
    \item what fields should be included in the output;
    \item how to phrase the constructed question, answer, and reasoning;
    \item whether the selected supervision is useful for downstream model training.
\end{itemize}

For example, given a game-like video of a player approaching a spear-wielding enemy, a valid model output may transform the clip into a predictive world-modeling sample. The model can select the frames where the player closes distance as input evidence and the frames where the enemy initiates a spear attack as output evidence:
\begin{verbatim}
{
  "task_type": "action anticipation / predictive world modeling",
  "question": "If the player keeps moving toward the spear-wielding enemy,
               what action is likely to happen next?",
  "answer": null,
  "cot": "The media is best suited for action-anticipation data rather than
          generic captioning because it contains a clear before-and-after
          temporal structure. Frames 8--12 show the informative input segment:
          the player is running down a rocky path, closing the distance to an
          enemy who is holding a spear and preparing to react. Frames 13--18
          show the target future segment: the enemy initiates a spear attack
          once the player enters striking distance. The constructed question
          turns this transition into a future-prediction task. The answer is
          null because the supervision is provided by the selected output
          frames. This example is useful for downstream training because it
          teaches a model to infer future combat behavior from approach
          direction, distance, enemy pose, and weapon state.",
  "field_design": {
    "task_type": "records the inferred data-construction objective",
    "question": "phrases the supervision as a conditional future-prediction
                 query",
    "answer": "kept null because the expected target is visual",
    "cot": "explains the task choice, evidence selection, and training value",
    "input_image": "contains the pre-event visual context",
    "output_image": "contains the future event used as supervision"
  },
  "input_image": [8, 9, 10, 11, 12],
  "output_image": [13, 14, 15, 16, 17, 18]
}
\end{verbatim}

Importantly, $\text{DataClaw}_0$-Intent does not have a unique ground-truth output. The same raw video could plausibly be converted into different useful training instances, such as future-frame prediction, action anticipation, event segmentation, or visual reasoning. Therefore, we do not score $\text{DataClaw}_0$-Intent with exact-match metrics against a single reference. Instead, the final evaluation is conducted through a user study. 100 human raters judge whether the model's constructed sample is useful, grounded in the media, appropriately structured, and aligned with a reasonable inferred data-construction goal.

\paragraph{Study protocol.} Raters were graduate students and research engineers with prior experience in multimodal data annotation, recruited internally through a laboratory mailing list; participation was voluntary and raters were free to stop at any point. Each rater was shown a description of the task and of how their responses would be used before beginning, and gave consent to that use; no personally identifying information was collected, and responses are stored in aggregate only. Rating took approximately 30 minutes per participant and was compensated at a rate consistent with local research-assistant pay.

Each of the 120 samples in the fuzzy-intent subset was judged independently by five raters along the four dimensions above, each on a five-point Likert scale; the 600 resulting judgements were spread over 100 raters, six samples each. We report the mean across raters and dimensions, after discarding submissions that failed two embedded attention checks (4.5\% of submissions). Inter-rater agreement measured by Krippendorff's $\alpha$ is 0.71, which is adequate for a subjective judgement of this kind but low enough that we treat the resulting scores as indicative rather than as a precise ranking. The verbatim instructions given to raters are reproduced in the supplementary material.

\subsubsection{Metrics}
\label{app:evaluation_metrics}

Structured multimodal data cannot be faithfully evaluated by conventional text-generation metrics such as BLEU or ROUGE. We therefore define a hierarchical score:
\begin{equation}
S_{total}=G_{json}\cdot
(\lambda_1S_{field}+\lambda_2S_{semantic}+\lambda_3S_{traj}),
\end{equation}
where \(G_{json}\in\{0,1\}\) is a hard JSON-validity gate, assigning zero score to malformed outputs. We use \(\lambda_1=0.20\), \(\lambda_2=0.35\), and \(\lambda_3=0.45\).

The field score measures schema integrity:
\begin{equation}
S_{field}=0.7C_{field}+0.3(1-R_{field}),
\end{equation}
where \(C_{field}\) is ground-truth field coverage and \(R_{field}\) is the truncated extraneous-field ratio, with robust key matching to tolerate minor formatting differences.

The semantic score measures embedding-based cosine similarity over textual fields:
\begin{equation}
S_{semantic}
=
\frac{
\alpha sim_{question}+\beta sim_{answer}+\gamma sim_{cot}
}{
\alpha+\beta+\gamma
},
\end{equation}
where \(\alpha=0.40\), \(\beta=0.40\), and \(\gamma=0.20\) by default, dynamically renormalized based on field availability.


For spatio-temporal samples, the visual score is defined based on trajectory-shape similarity:
\begin{equation}
S_{traj}
=
\exp(-d_{shape}/\tau_s),
\end{equation}
where trajectories are normalized and resampled to \(K=50\); \(d_{shape}\) denotes the trajectory Mean Absolute Error (MAE), and the temperature hyperparameter is set to \(\tau_s = 0.10\).

\section{Training and Deployment Details}
\label{app:training_details}

This section provides details about GRPO training, reward design, and the Omni/Expert deployment paradigms.

\subsection{Rule-Driven Reinforcement Learning via GRPO}
\label{app:grpo}
To overcome hallucinations and formatting collapses in long-sequence multimodal tasks, we introduce reinforcement learning atop the initial policy \(\pi_{SFT}\). Traditional multimodal RLHF requires training a reward model of comparable size to the policy model, which incurs prohibitive GPU memory overhead when processing long videos or high-resolution image sequences. Therefore, $\text{DataClaw}_0$ employs Group Relative Policy Optimization (GRPO), eliminating the reliance on a Critic model through intra-group relative advantage estimation.

Crucially, because $\text{DataClaw}_0$'s objective is to generate strictly structured data tailored to specific intents, we can entirely discard subjective and computationally expensive neural network reward models in favor of model-free, deterministic \textit{Rule-based Rewards}. For any trajectory \(Y\) generated by the model, we define the joint reward function \(R(Y)\):
\begin{equation}
R(Y) = \lambda_1 R_{format}(Y, \Phi) + \lambda_2 R_{anchor}(Y, A) + \lambda_3 R_{eff}(Y)
\end{equation}

\begin{itemize}[leftmargin=20pt, labelsep=4pt]
    \item \textbf{Format Compliance Reward (\(R_{format}\)):} We utilize an Abstract Syntax Tree (AST) or regex parser to strictly verify whether the generated content \(Y\) is 100\% compliant with the target schema \(\Phi\). Successful parsing yields a high positive reward, whereas structural or syntactic errors result in an immediate truncation penalty.
    

\item \textbf{Spatio-Temporal Anchor Reward (\(R_{anchor}\)):} 
This reward is designed for embodied and video-based tasks that exhibit complex spatio-temporal dynamics. 
It measures the \textit{temporal shape similarity} between the predicted and ground-truth trajectories—defined as the similarity of their frame-wise distributions in the constructed data sequence. 
To ensure temporal alignment across trajectories of varying lengths, we first normalize discrete frame indices to a standardized interval and apply interpolation to resample each trajectory into \(K = 50\) uniformly distributed alignment points. 
We then compute the Mean Absolute Error (MAE) between the predicted and ground-truth distributions of these alignment points to obtain the trajectory deviation \(d_{shape}\). 
Finally, this deviation is converted into a similarity-based reward through an exponential mapping:
\begin{equation}
R_{anchor} = \exp\left(-\frac{d_{shape}}{\tau_s}\right)
\end{equation}
where the temperature hyperparameter \(\tau_s = 0.10\) provides tolerance to minor misalignments. 
This design allows $\text{DataClaw}_0$ to receive fine-grained quantitative feedback reflecting both the temporal shape coherence and the overall spatio-temporal alignment quality.


\item \textbf{Reasoning Efficiency Penalty (\(R_{eff}\)):} 
To prevent the model from producing overly verbose or trivially short Chain-of-Thought (CoT) reasoning, we introduce a dynamic length regularization mechanism. 
Since the appropriate CoT length varies with the underlying sample complexity, an ideal reasoning trace should be sufficiently elaborate to cover key analytical steps, yet concise enough to avoid redundant self-expansion. 
To operationalize this notion, we assess CoT length adaptively within each GRPO rollout group. 
Specifically, we identify the rollouts whose overall reward ranks within the top 50\% of the group and treat their CoT lengths as reference standards. 
For the remaining rollouts, CoT sequences that are significantly longer or shorter than this reference are penalized, while those falling close to the reference range receive a slight positive adjustment. 
This relative, group-wise normalization encourages the policy to learn reasoning behaviors that are both efficient and well-calibrated to task complexity.
    
\end{itemize}

\textbf{GRPO Optimization Objective:}
During training, for a given input \((X_{raw}, I)\), the current policy model \(\pi_\theta\) samples a candidate group \(G\) containing \(G\) diverse outputs: \(G = \{Y^{(1)}, Y^{(2)}, \dots, Y^{(G)}\}\).
We compute the joint reward \(R^{(g)} = R(Y^{(g)})\) for each candidate and normalize them within the group to obtain the relative advantage estimation:
\begin{equation}
\hat{A}^{(g)} = \frac{R^{(g)} - \mu_R}{\sigma_R}
\end{equation}
where \(\mu_R\) and \(\sigma_R\) are the mean and standard deviation of the \(G\) rewards in the group, respectively. Subsequently, we update the model parameters \(\theta\) by maximizing the following GRPO objective function:
\begin{equation}
J(\theta) = \mathbb{E}\left[\frac{1}{G}\sum\limits_{g=1}^G \left(\min\left(\rho^{(g)}(\theta)\hat{A}^{(g)}, \text{clip}(\rho^{(g)}(\theta), 1-\epsilon, 1+\epsilon)\hat{A}^{(g)}\right) - \beta D_{KL}(\pi_\theta \parallel \pi_{ref})\right)\right]
\end{equation}
where the importance sampling ratio is \(\rho^{(g)}(\theta) = \frac{\pi_\theta(Y^{(g)}\mid X_{raw},I)}{\pi_{old}(Y^{(g)}\mid X_{raw},I)}\), and \(\epsilon\) is the clipping hyperparameter. 

Through this ``SFT Initialization + Rule-Reward Driven GRPO'' paradigm, $\text{DataClaw}_0$ significantly stimulates the model's agentic capability to proactively tailor high-value structured knowledge from complex multimodal streams, particularly excelling in spatio-temporal grounding due to the injection of \(R_{anchor}\).

\subsection{Inference Architecture \& Deployment Paradigms}
\label{app:deployment}

To transform the GRPO-aligned policy model \(\pi_\theta\) into a practical structured tailoring agent, we design an inference architecture that emphasizes output validity, grounding reliability, and deployment flexibility. Given raw multimodal streams \(X_{raw}\) and user-defined intents \(I\), $\text{DataClaw}_0$ generates schema-aligned structured outputs \(Y_{struct}\).

\subsubsection{Core System Architecture}
\label{app:inf}
$\text{DataClaw}_0$'s inference process consists of three decoupled modules:
    
    

\begin{itemize}[leftmargin=20pt, labelsep=4pt]
    \item \textbf{Multimodal Ingestion \& Intent Parsing:} This module receives raw multimodal data streams \(X_{raw}\) and user-defined intents \(I\), and converts them into structured context sequences readable by the policy model. 
    In the current implementation, we employ rule-based automation scripts to segment long video trajectories into broad yet semantically coherent clips that contain key events based on raw dataset annotations. 
    This step serves as a scalable preprocessing pipeline for multimodal grounding. 
    In future iterations, this stage will be extended into a more intelligent pipeline using a long-video understanding model capable of adaptively identifying event boundaries and intent-relevant sub-sequences.
    
    \item \textbf{Schema-Constrained Inference Engine:} This module deploys the policy model \(\pi_\theta\) to generate structured outputs. To reduce formatting failures, a lightweight schema-aware constraint can be applied during decoding, suppressing invalid continuations that violate the target schema \(\Phi\).
    
    \item \textbf{Structural Verification \& Grounding:} 
This module conducts post-hoc structural and semantic verification by leveraging factual anchors \(A\) together with quantitative evaluation metrics. 
When the generated output \(Y_{struct}\) contains concrete elements such as UI components, spatial coordinates, timestamps, or action trajectories, the system cross-checks these elements against extracted anchors to ensure factual consistency and spatial grounding. 
Furthermore, the verification process is quantitatively guided by three complementary metrics: 
\textit{(i)} schema–field correctness for structural validity, 
\textit{(ii)} textual semantic alignment for content faithfulness, and 
\textit{(iii)} trajectory–shape similarity for spatio-temporal coherence. 
Together, these mechanisms ensure that the tailored outputs are both semantically grounded and structurally reliable.
\end{itemize}

\subsubsection{Omni vs. Expert Deployment Paradigms}

Considering the heterogeneity of multimodal data and varying deployment requirements, $\text{DataClaw}_0$ supports two complementary architectures:
\begin{itemize}[leftmargin=20pt, labelsep=4pt]
    \item \textbf{$\text{DataClaw}_0$-O (Omni Tailoring Agent):} A unified model jointly trained across all domains. It serves as a generalist agent for diverse cross-domain intents \(I\) within a single model.
    
    \item \textbf{$\text{DataClaw}_0$-E (Expert Tailoring System):} A decoupled architecture composed of domain-specific tailoring agents, such as GUI, embodied AI experts. In deployment, the corresponding expert is selected according to the target scenario, domain setting, or user-specified configuration. This paradigm improves domain-specific robustness and allows modular updates.
\end{itemize}

\section{Downstream Evaluation}
\label{app:downstream_protocols}

\subsection{Detailed Analysis of Downstream Application}
\label{app:downstream}

The downstream evaluation aims to verify whether the structured data generated by $\text{DataClaw}_0$ can provide effective supervision for targeted model refinement. We consider three representative multimodal tasks: long-horizon GUI navigation~\cite{wang2025opencua}, action video generation~\cite{grauman2022ego4d}, and spatio-temporal VQA~\cite{remot}. These tasks cover different output spaces and reasoning requirements, including action planning, visual dynamics generation, and fine-grained temporal understanding.

For a fair comparison, all training sets are constructed from the same raw input streams. Specifically, each raw stream is processed by three different annotators: the base model itself, Gemini-3.1-Pro-Preview, and $\text{DataClaw}_0$. The base-model processing setting serves as a self-refinement baseline, while Gemini represents a strong proprietary-model annotator. To ensure that the comparison focuses on data quality rather than data quantity, we apply the same coarse rule-based filtering procedure to remove malformed samples and then randomly sample an equal number of valid instances from each processed data pool.

As shown in Table~\ref{tab:downstream_results}, fine-tuning on self-processed data only provides limited improvements over the zero-shot base models, suggesting that simple self-refinement is insufficient for these challenging multimodal tasks. In contrast, both Gemini-processed and $\text{DataClaw}_0$-processed data lead to substantial gains, demonstrating that high-quality structured supervision is crucial for downstream adaptation.

A closer comparison between Gemini and $\text{DataClaw}_0$ shows a consistent trade-off. Gemini achieves slightly better results on some partial or step-level metrics, such as GUI SSR and VQA Partial Accuracy. This suggests that strong proprietary models can provide rich intermediate supervision that benefits local or partially correct predictions. However, $\text{DataClaw}_0$ achieves stronger results on several end-to-end metrics that better reflect final task completion. For example, $\text{DataClaw}_0$ obtains higher GUI TSR and VQA Overall Accuracy, indicating better transfer to complete task-solving ability. In action video generation, $\text{DataClaw}_0$ also achieves lower FVD and higher Contact mAP, showing improved visual quality and stronger modeling of physical interactions.

These results support the motivation of $\text{DataClaw}_0$. Instead of producing verbose general-purpose annotations, $\text{DataClaw}_0$ is optimized to extract compact, schema-aligned, and task-relevant structured supervision from multimodal streams. Under the same data budget, such targeted supervision can be more effective for improving end-to-end performance. Therefore, $\text{DataClaw}_0$ provides a practical and controllable alternative to proprietary annotators for downstream multimodal data construction.

\subsection{Unified Refinement Protocol}
\label{app:unified_downstream_protocol}

For each downstream task, we use the same raw data streams and process them with different annotators: the base model itself, a strong closed-source VLM, and $\text{DataClaw}_0$. All annotators receive the same task intent and target schema. Their outputs are passed through the same coarse filtering pipeline. We then sample an equal number of valid instances for downstream fine-tuning.

The protocol is:

\begin{enumerate}[leftmargin=*]
    \item Select raw multimodal streams for a downstream task.
    \item Process the same raw streams using Self-Refinement, Gemini-3.1-Pro / Qwen3.5, and $\text{DataClaw}_0$.
    \item Apply the same schema and quality filters.
    \item Randomly sample the same number of valid instances from each method.
    \item Fine-tune the same downstream base model using identical hyperparameters.
    \item Evaluate on held-out task-specific benchmarks.
\end{enumerate}



\subsection{GUI Navigation}
\label{app:gui_downstream}

\paragraph{Downstream model.}
We fine-tune Qwen3.5-4B. The model predicts actions from screenshots, task instructions, and interaction history.

\paragraph{Metrics.}
We report Step Success Rate (SSR) and Task Success Rate (TSR). SSR measures the fraction of predicted steps that match the reference action or lead to the correct UI state. TSR measures whether the full task is completed successfully within a maximum number of steps. Formally,
\begin{equation}
\mathrm{SSR}
=
\frac{\#\text{successful steps}}{\#\text{total steps}},
\quad
\mathrm{TSR}
=
\frac{\#\text{successful tasks}}{\#\text{total tasks}}.
\end{equation}

\begin{figure}[h]
\centering
\includegraphics[width=\linewidth]{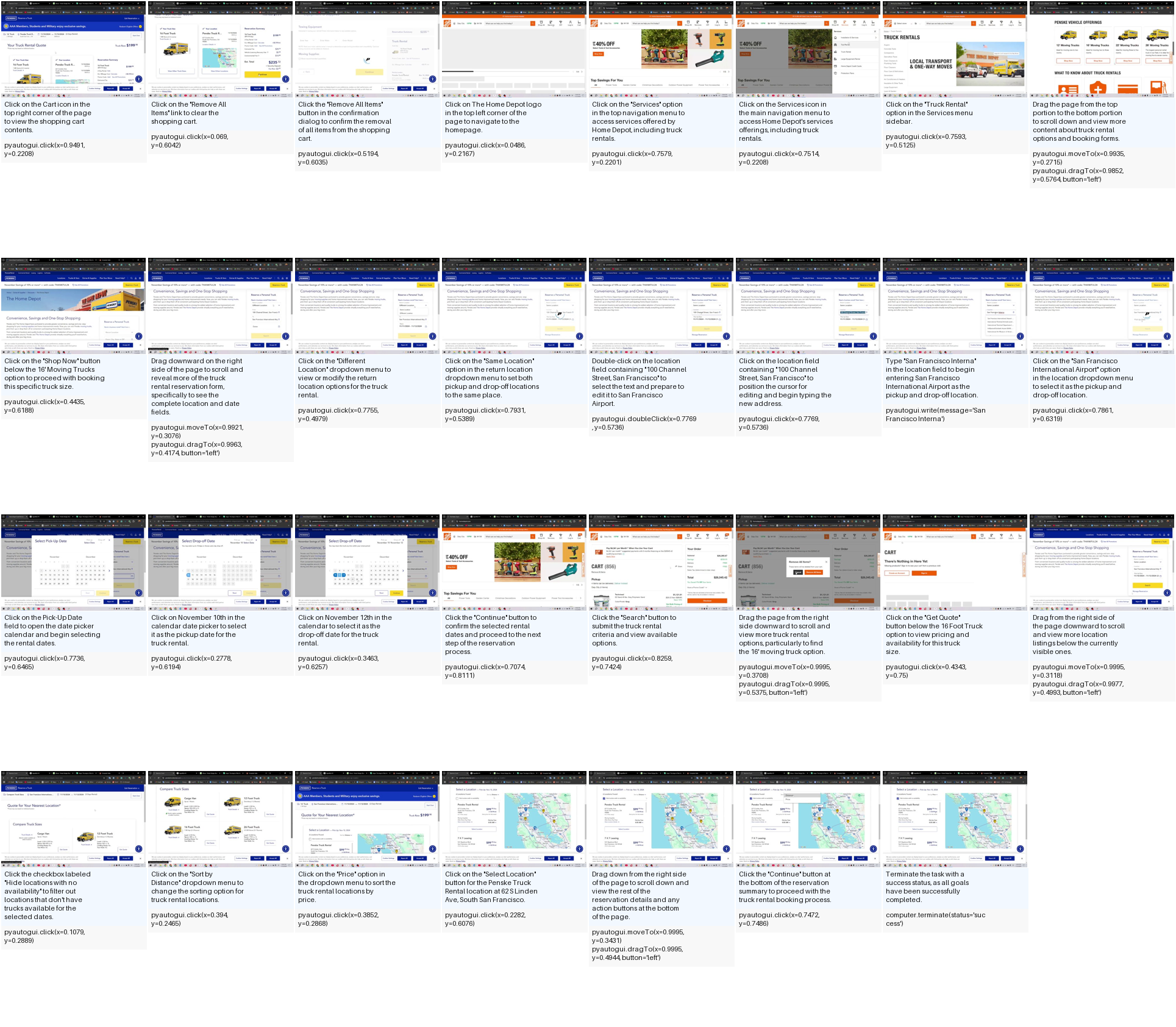}
\caption{Qualitative example for GUI navigation.}
\label{fig:gui_visualization}
\end{figure}

Figure~\ref{fig:gui_visualization} visualizes a representative downstream example.

\subsection{Action Video Generation}
\label{app:video_downstream}

\paragraph{Downstream model.}
We fine-tune Wan2.2-I2V-5B. The model receives an input image and text prompt and generates a short video.

\paragraph{Fine-tuning hyperparameters.}
Table~\ref{tab:video_hyperparams} lists the video generation fine-tuning setup.

\begin{table}[h]
\centering
\small
\caption{Video generation downstream fine-tuning hyperparameters.}
\label{tab:video_hyperparams}
\begin{tabular}{ll}
\toprule
Hyperparameter & Value \\
\midrule
Base model & Wan2.2-I2V-5B \\
Training clips & 200 \\
Resolution & $480 \times 832$ \\
Frames per clip & 81 \\
Epochs & 3  \\
Learning rate & $1 \times 10^{-5}$ \\
Batch size & 1 \\
Hardware & 8 $\times$ NVIDIA A100 80GB \\
\bottomrule
\end{tabular}
\end{table}

\begin{figure}[h]
\centering
\includegraphics[width=\linewidth]{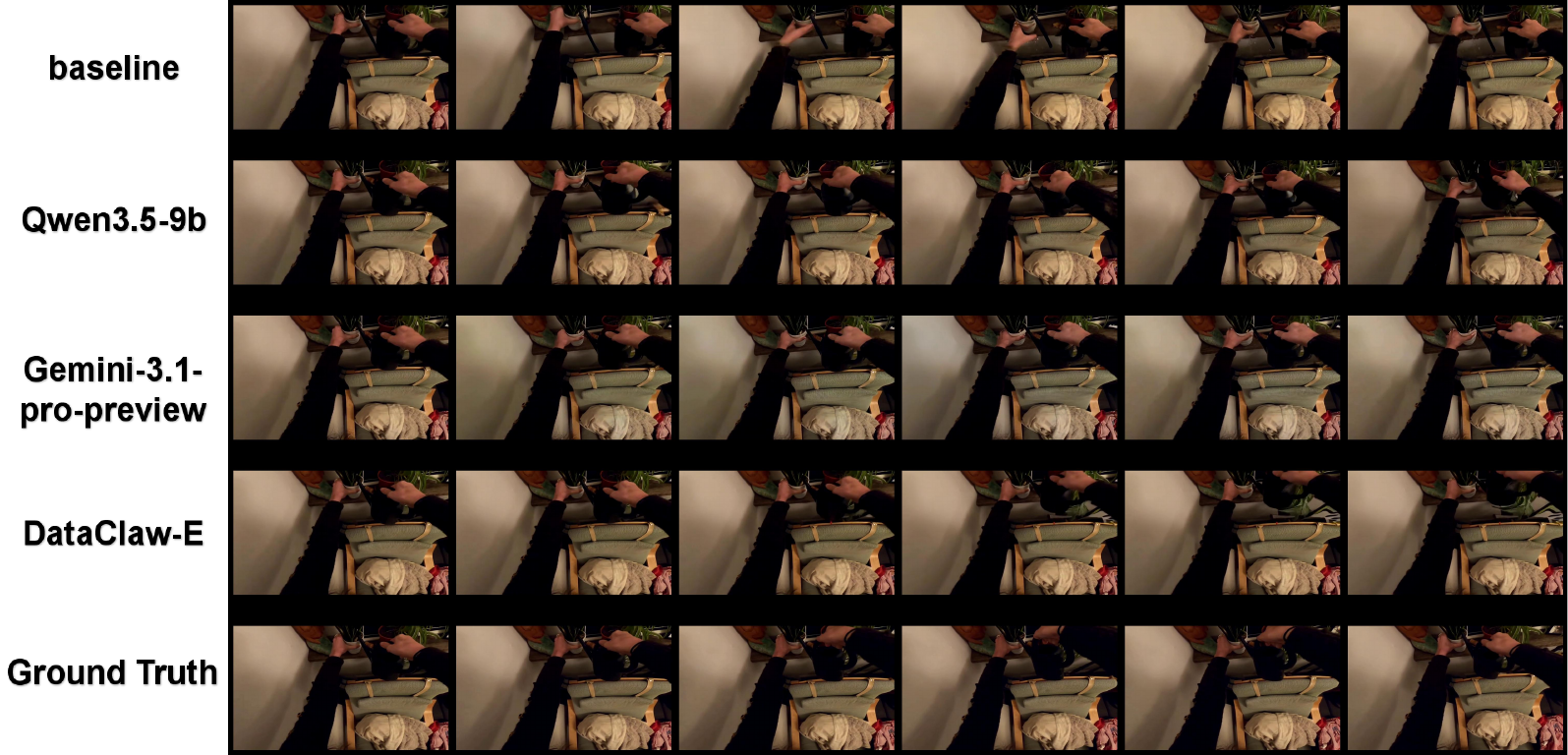}
\caption{Qualitative comparison for action video generation.}
\label{fig:video_visualization}
\end{figure}

\paragraph{Visualization.}
Figure~\ref{fig:video_visualization} compares videos generated by models trained with different refined datasets.

\subsection{Spatio-temporal VQA}
\label{app:vqa_downstream}

\paragraph{Downstream model.}
We fine-tune Qwen3.5-4B. The model receives video frames and a question and predicts the answer.

\paragraph{Metrics.}
We report Partial Accuracy and Overall Accuracy. Partial Accuracy gives credit for partially correct structured answers, such as correctly identifying the object but missing the temporal order. Overall Accuracy requires the full answer to match the reference. 



\begin{figure}[h]
\centering
\includegraphics[width=\linewidth]{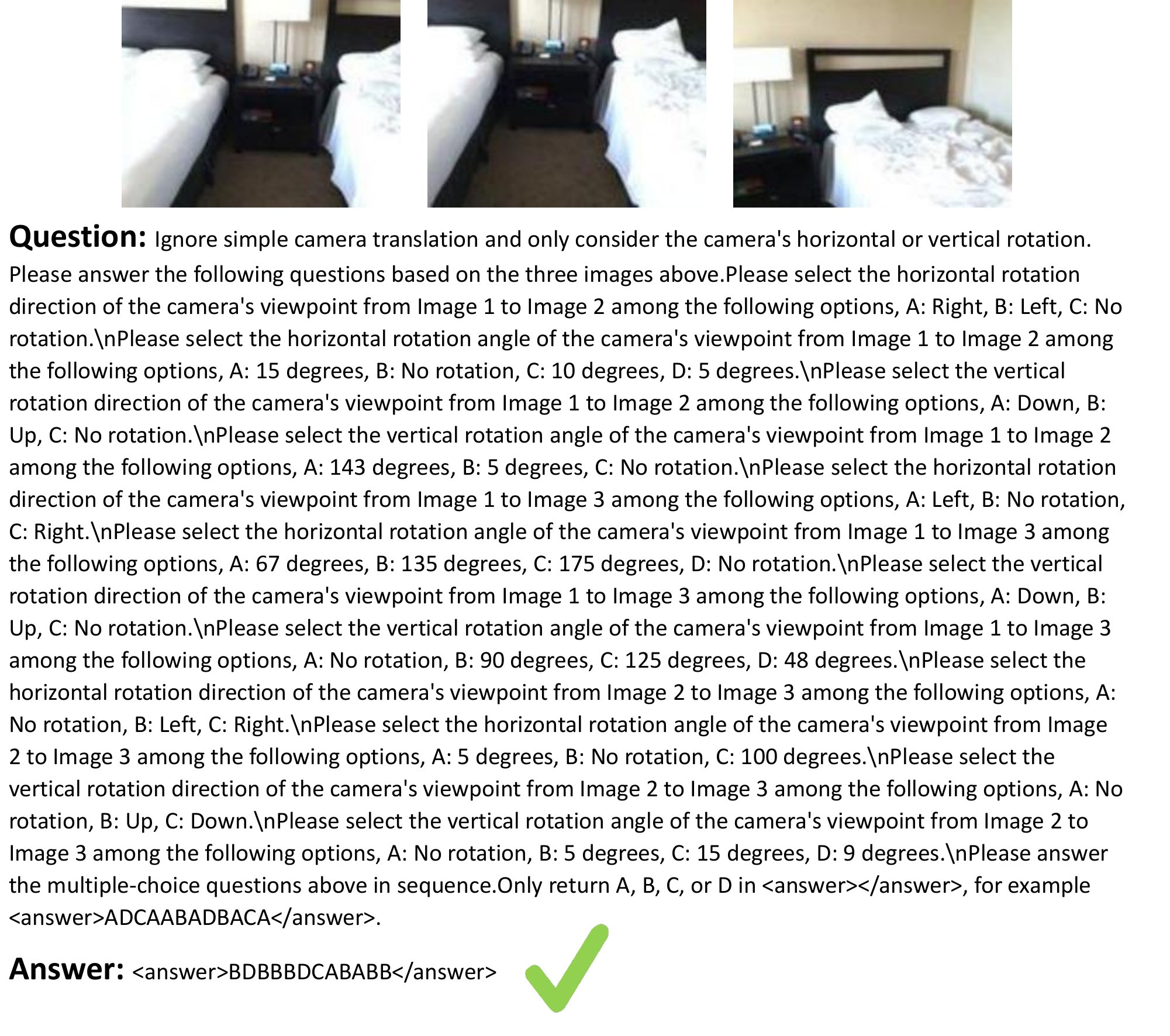}
\caption{Qualitative example for spatio-temporal VQA.}
\label{fig:vqa_visualization}
\end{figure}

\paragraph{Visualization.}
Figure~\ref{fig:vqa_visualization} shows a representative VQA example.

\section{Additional Experiments and Ablations}
\label{app:additional_experiments}

\subsection{Scaling Laws and Emergent Diversity}
\label{app:scaling_diversity}

In this section we investigate the training dynamics of our framework: when a domain-decoupled expert architecture beats a unified model and when it does not, and how the generalization behaviour of $\text{DataClaw}_0$ changes with backbone capacity.

\textbf{Task interference at 9B.} At the 9B scale the unified model \textbf{$\text{DataClaw}_0$-O} does not exhibit a smooth, log-linear scaling curve. Its overall score starts at 53.60 with 1/15 of the corpus, drops to 47.23 at 7/15, rebounds to 57.84 at 12/15, and oscillates thereafter, whereas the \textbf{$\text{DataClaw}_0$-E} ensemble scales monotonically to a combined 68.86. We attribute the instability to task interference in the shared weights of a 9B backbone: the five extraction tasks span very different distributions, and forcing one compact model to satisfy all of them simultaneously produces gradient conflict. Routing to domain-specific experts avoids the conflict by letting each model fit its own local distribution.

\textbf{Why this does not generalise to larger backbones.} Read on its own, the 9B result would justify expert routing as the correct architecture. The 4B and 27B runs show that it is instead a remedy for limited capacity. At 4B the interference is worse (the Expert ensemble leads by 14.6 points of Overall); at 9B it is 10.5; at 27B the ordering reverses and the unified model leads by 5.9, with a monotone rather than oscillating scaling curve (Sec.~\ref{subsec:scaling_and_diversity}). The instability documented above is therefore a symptom of the backbone being too small to hold five domains at once, not evidence that the domains are fundamentally incompatible. We retain the 9B analysis because it is the regime most groups will actually operate in, and because the contrast between the two regimes is what identifies the mechanism: interference and transfer are the same interaction seen from opposite sides of a capacity threshold.

\textbf{Emergent Diversity and Intent Comprehension.}
To examine whether $\text{DataClaw}_0$ merely memorizes surface patterns or demonstrates genuine generalization ability, 
we extract the semantic embeddings of three data sources: the original Raw Data, 
the data refined by the base model (Qwen3.5-9B), and the data refined by $\text{DataClaw}_0$. 
\begin{figure}[t]
  \centering
  \includegraphics[width=0.78\linewidth]{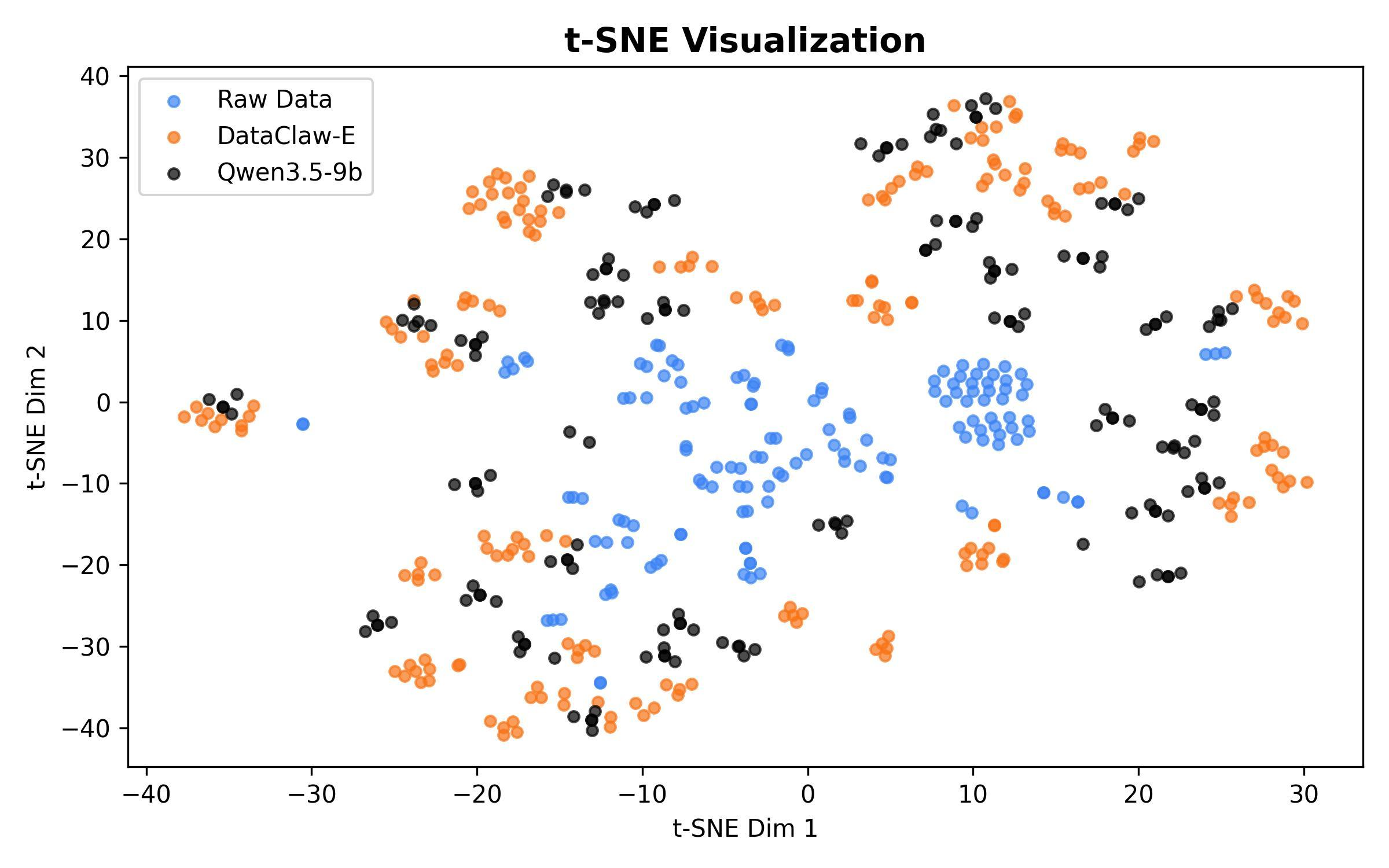}
  \caption{t-SNE of tailored outputs, the raw input distribution, and the untrained backbone. Produced with the 9B model; to be regenerated for the 27B configuration.}
  \label{fig:tsne_appendix}
\end{figure}

We visualize their feature distributions using t-SNE dimensionality reduction (Figure~\ref{fig:tsne_appendix}). 

The visualization shows that the feature space of the base model’s outputs expands slightly beyond that of the Raw Data, 
indicating limited diversification in its refinements. 
In contrast, the data refined by $\text{DataClaw}_0$ exhibits a substantially broader and more evenly distributed coverage across the semantic space. 
This suggests that $\text{DataClaw}_0$ injects stronger emergent diversity into the generated trajectories, 
revealing new clusters and long-tail patterns that the base model fails to capture. 
Rather than marginally shifting the training distribution, 
$\text{DataClaw}_0$ reconstructs it into a richer and more heterogeneous semantic landscape.

To further quantify this capability, we build a \textbf{$\text{DataClaw}_0$-Intent} evaluation subset composed entirely of vague, high-level instructions. 
As shown in Table~\ref{tab:main_results}, $\text{DataClaw}_0$ significantly surpasses its base model (Qwen3.5-9B) 
and approaches the performance of Gemini 3.1-pro-preview on this challenging fuzzy-instruction benchmark.

\section{Qualitative Analysis}
\label{app:qualitative_limitations}

\subsection{Successful Case Studies}
\label{app:success_cases}

We showcase qualitative case studies covering representative domains to exemplify $\text{DataClaw}_0$’s end-to-end process.
Each case presents the raw input, user intent, and $\text{DataClaw}_0$ output.




\begin{figure}[h]
\centering
\includegraphics[width=\linewidth]{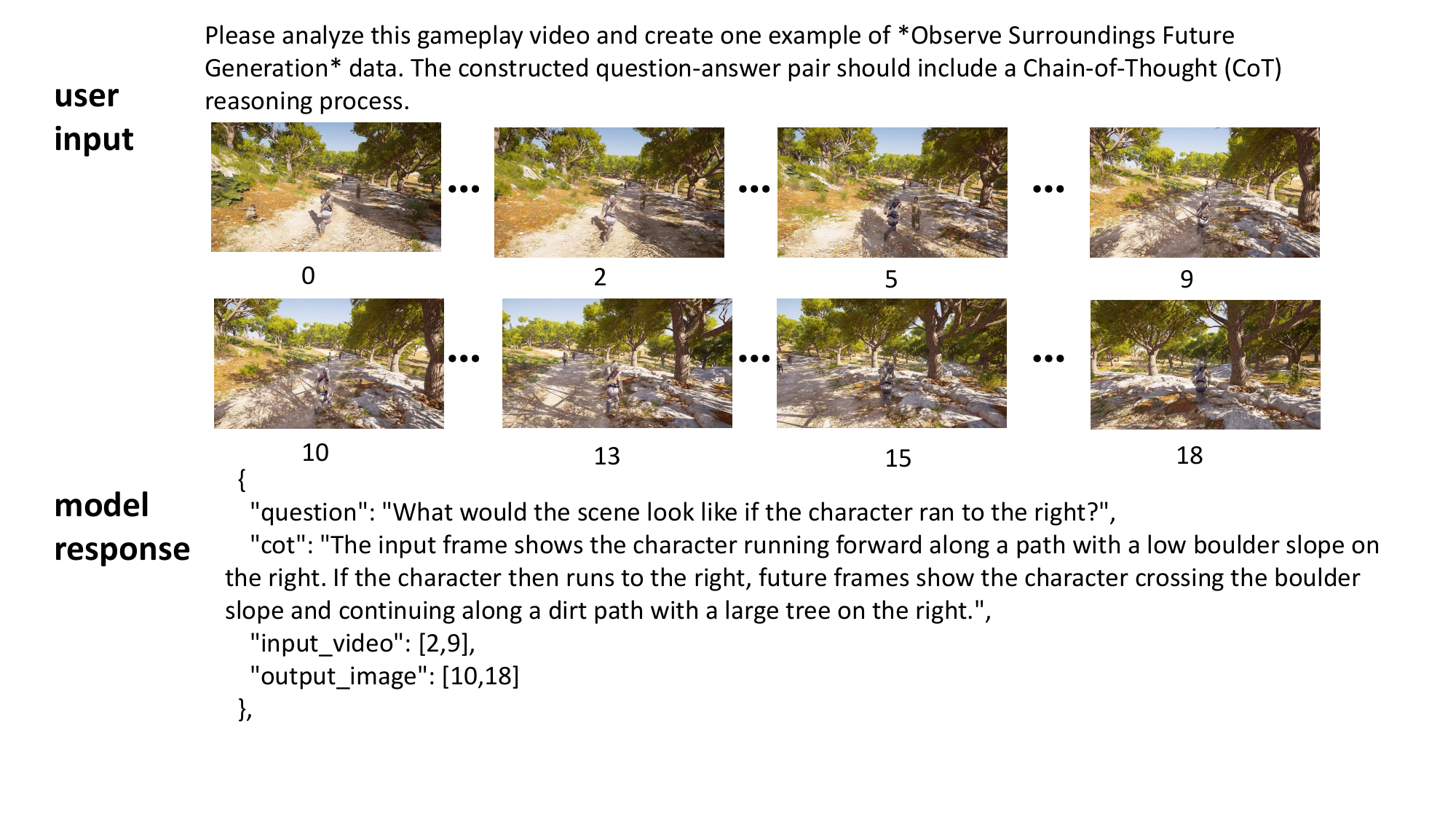}
\caption{Successful world-model or video-generation tailoring case.}
\label{fig:aigc_success_case}
\end{figure}

\begin{figure}[h]
\centering
\includegraphics[width=\linewidth]{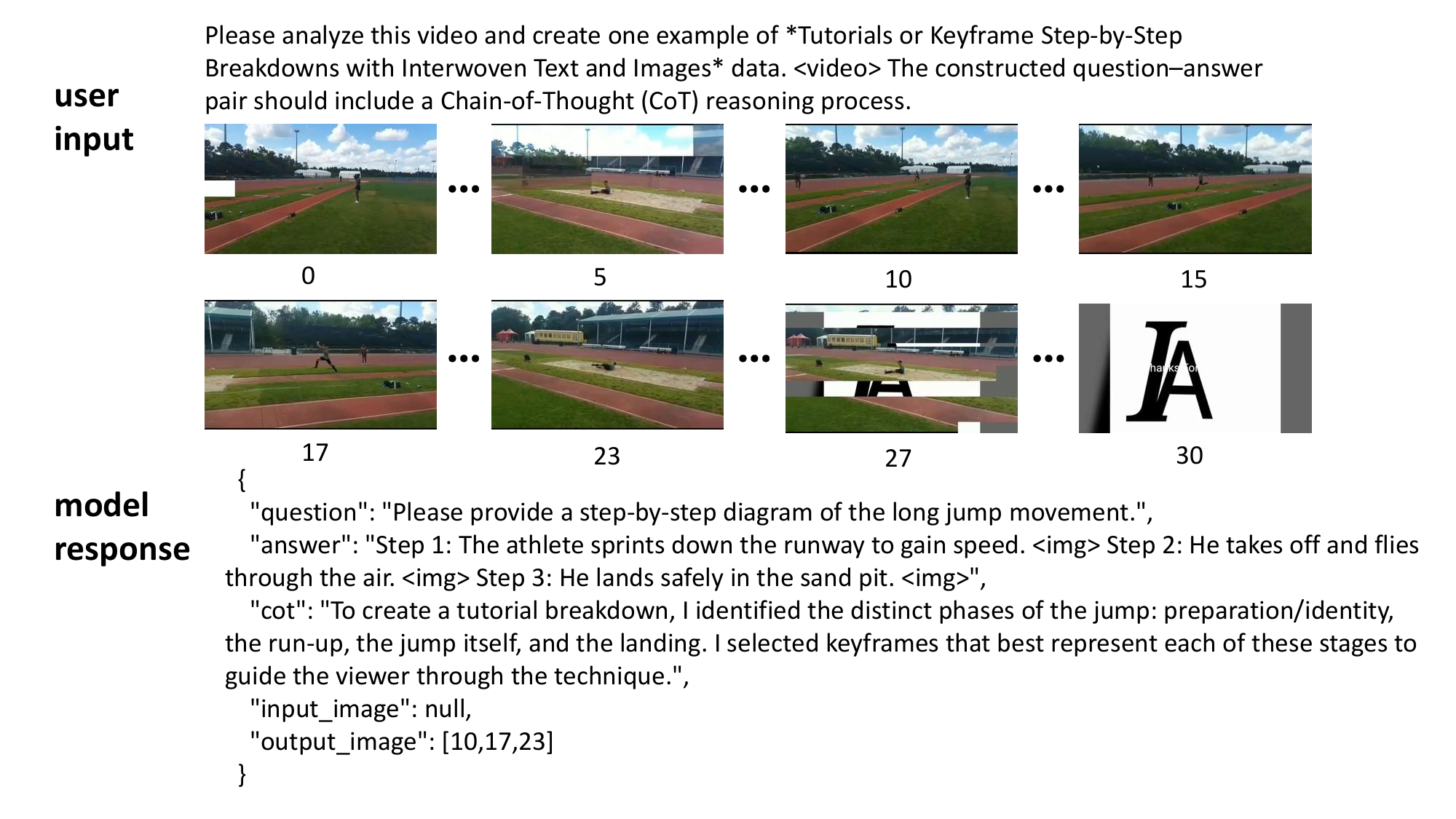}
\caption{Successful daily-life tailoring case.}
\label{fig:daily_success_case}
\end{figure}

\subsection{Failure Cases}
\label{app:failure_cases}

\begin{figure}[h]
\centering
\includegraphics[width=\linewidth]{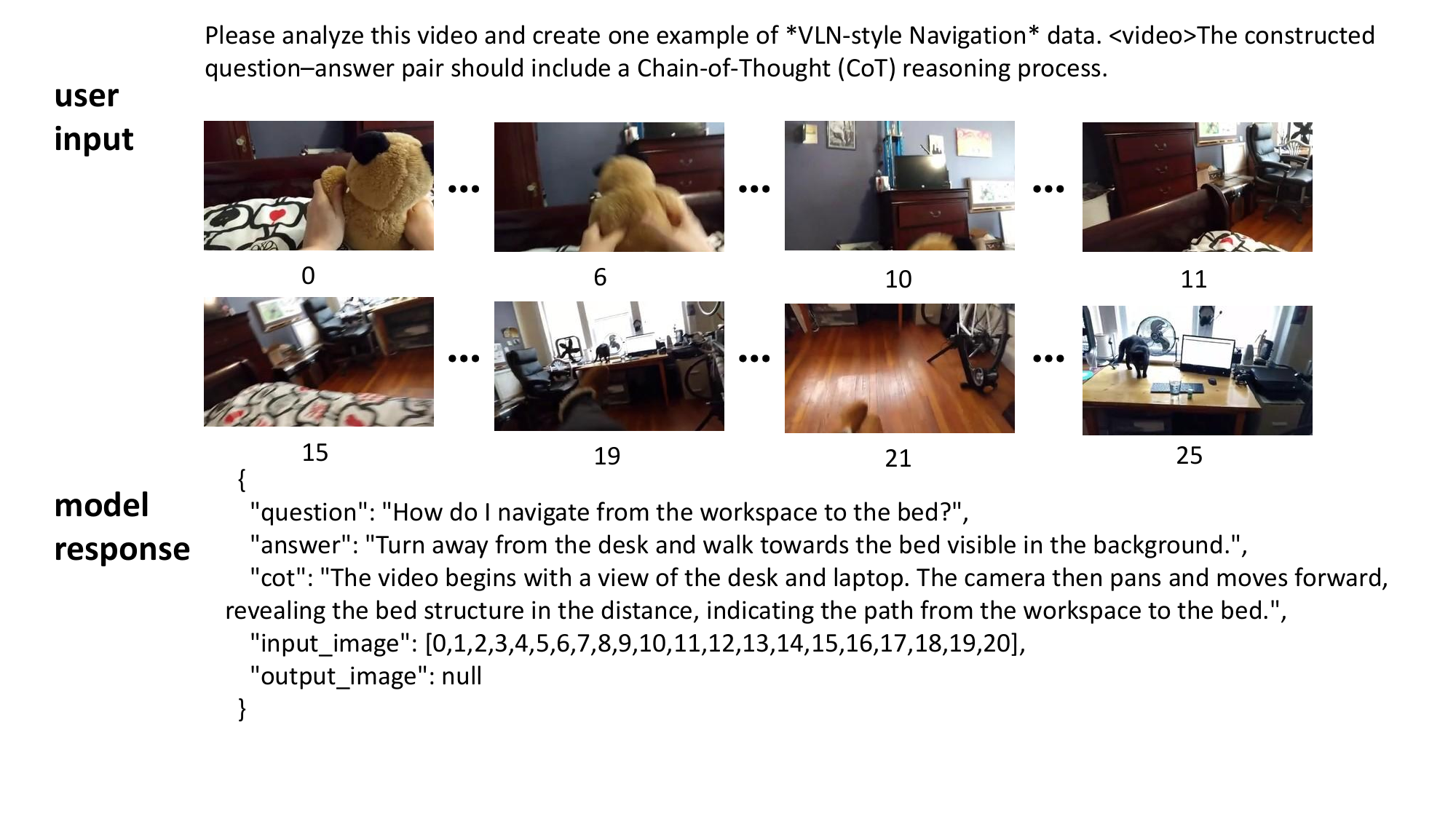}
\caption{Failed daily-life tailoring case.}
\label{fig:daily_failure_case}
\end{figure}

Although $\text{DataClaw}_0$ improves structured multimodal data tailoring, it still fails in several scenarios. 
As shown in Figure~\ref{fig:daily_failure_case}, the constructed sample is semantically correct---the question asks how to navigate from the workspace to the bed, 
and the generated answer properly describes that transition. 
However, the CoT text describes the video as starting from the workspace and then moving toward the bed, 
while the actual input frames (\(0 \rightarrow 20\)) are temporally ordered in the opposite direction (from the bed to the workspace). 

Such temporal inconsistencies are occasionally observed when large multimodal language models generate narrative-style reasoning. 
They tend to infer a likely or contextually coherent event flow based on spatial cues rather than the exact chronological order of frames, 
a behavior commonly referred to as \textit{temporal hallucination}. 
This phenomenon is largely inherent to the base model’s reasoning prior.
Although minor, these cases highlight the intrinsic difficulty of enforcing strict temporal grounding in long-horizon multimodal reasoning, 
which remains an open challenge for current LLM-based architectures.



\section{Amortised Cost Analysis}
\label{app:cost}

A tailoring model distilled from a proprietary annotator is only worth building if it is cheaper than calling that annotator directly. Because the two options have different cost structures -- one is dominated by a fixed up-front investment, the other is billed per sample -- their comparison is not a single number but a break-even volume. This section makes that comparison explicit.

\paragraph{Cost model.} Let $c_{\mathrm{api}}$ denote the price of annotating one sample through a hosted API, and let $c_{\mathrm{local}}$ denote the marginal cost of producing one sample with a self-hosted $\text{DataClaw}_0$-9B, i.e.\ accelerator rental divided by serving throughput. Acquiring the tailoring capability incurs a one-off cost
\begin{equation}
C_{\mathrm{fixed}} = \underbrace{N_{\mathrm{train}}\, c_{\mathrm{teacher}}}_{\text{teacher calls for the SFT corpus}} \;+\; \underbrace{(T_{\mathrm{sft}} + T_{\mathrm{grpo}})\, r_{\mathrm{gpu}}}_{\text{training compute}},
\end{equation}
where $N_{\mathrm{train}}$ is the size of the distilled corpus, $T$ denotes accelerator-hours and $r_{\mathrm{gpu}}$ the hourly rate. Producing $n$ tailored samples then costs $C_{\mathrm{fixed}} + n\,c_{\mathrm{local}}$ for the self-hosted route against $n\,c_{\mathrm{api}}$ for the API route, and the two are equal at
\begin{equation}
n^{\star} = \frac{C_{\mathrm{fixed}}}{c_{\mathrm{api}} - c_{\mathrm{local}}} .
\label{eq:breakeven}
\end{equation}
Beyond $n^{\star}$ the self-hosted route is strictly cheaper, and the gap widens linearly in $n$.

\paragraph{Instantiation.} We instantiate Eq.~\ref{eq:breakeven} against two hosted annotators at list prices current at the time of writing: Gemini-3.1-Pro, the teacher used to build our corpus, and Kimi-K2.6 as a lower-priced alternative. Table~\ref{tab:cost_assumptions} states every quantity the estimate depends on. We report these as engineering estimates derived from the stated assumptions rather than as instrumented measurements, so that a reader who disagrees with any assumption can substitute their own value and recompute; the accompanying analysis script does exactly this.

\begin{table}[t]
\centering
\small
\caption{Quantities entering the cost estimate. API rates are list prices as of July 2026. The accelerator rate is the upper end of the cross-provider median band for rented A100-80GB capacity, matching the hardware used for all training reported in this paper, and is deliberately more conservative than marketplace floors near \$0.65/GPU-h. Training cost and serving throughput are given per backbone, since the two operating points differ mainly in these two rows.}
\label{tab:cost_assumptions}
\begin{tabular}{llrr}
\toprule
Group & Quantity & 9B & 27B \\
\midrule
\multirow{4}{*}{API list price}
 & Gemini-3.1-Pro, input / output (per 1M tokens) & \multicolumn{2}{c}{\$2.00 / \$12.00} \\
 & Kimi-K2.6, input / output (per 1M tokens) & \multicolumn{2}{c}{\$0.95 / \$4.00} \\
 & Input tokens per tailoring call ($\approx$24 frames + text) & \multicolumn{2}{c}{12{,}000} \\
 & Output tokens per tailoring call & \multicolumn{2}{c}{1{,}200} \\
\midrule
\multirow{4}{*}{One-off cost}
 & Retained examples in the distilled corpus & \multicolumn{2}{c}{34{,}700} \\
 & Teacher calls per retained example (validation yield) & \multicolumn{2}{c}{1.25} \\
 & SFT, 1 epoch & 96 GPU-h & 288 GPU-h \\
 & GRPO, $G{=}8$ rollouts & 288 GPU-h & 864 GPU-h \\
\midrule
\multirow{2}{*}{Serving}
 & Rented A100-80GB capacity & \multicolumn{2}{c}{\$1.80 / GPU-h} \\
 & Throughput, batched decoding & 200 / GPU-h & 67 / GPU-h \\
\bottomrule
\end{tabular}
\end{table}

\paragraph{Serving cost favours the joint configuration.} The two backbones differ in the direction one would expect: the 27B model is roughly three times more expensive to train and produces roughly one third as many samples per accelerator-hour. Its one-off cost is \$3{,}740 against \$2{,}360 for the 9B model, and its marginal cost per sample is \$0.027 against \$0.009. Taken alone this would make the larger backbone the worse economic choice.

It does not, because the configuration that wins on accuracy also removes a multiplier that the cost model above ignores. The sharded configuration requires five domain experts to be held resident to serve arbitrary intents, whereas the joint configuration requires one. At 27B a single joint model therefore occupies less accelerator memory than five 9B experts and needs no router, so the comparison that matters in deployment is one 27B model against five 9B models rather than one against one. We report the single-model figures in Table~\ref{tab:cost_assumptions} because they are the ones a reader can verify, and note that they understate the joint configuration's advantage.

\begin{figure}[t]
  \centering
  \includegraphics[width=\linewidth]{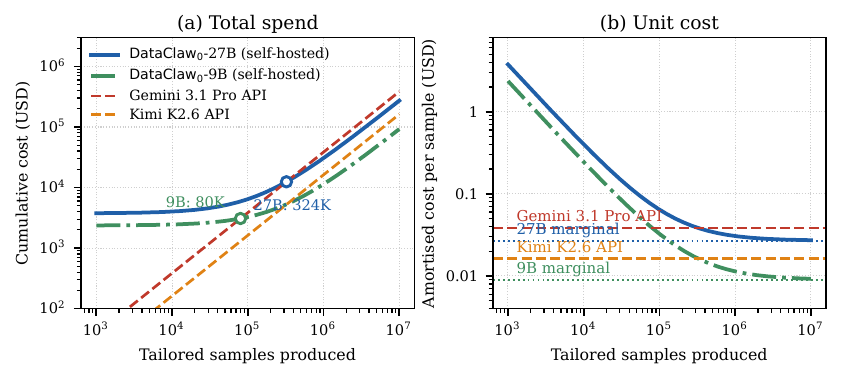}
  \caption{Amortised cost of producing tailored data, for both backbones. \textbf{(a)} Cumulative spend against the number of samples produced. Each self-hosted curve starts at the one-off cost of distilling the tailoring capability and then grows with the marginal cost of local inference, whereas API annotation grows linearly from the origin; circles mark the break-even volumes of Eq.~\ref{eq:breakeven} against the teacher. \textbf{(b)} The same data as cost per sample, decaying towards each backbone's marginal cost while the API rates stay flat. The 27B marginal cost sits just below Gemini-3.1-Pro but above Kimi-K2.6, which is why it never breaks even against the latter. Both axes are logarithmic.}
  \label{fig:cost_breakeven}
\end{figure}

\paragraph{Break-even.} Substituting into Eq.~\ref{eq:breakeven}, and using the per-sample price of Gemini-3.1-Pro at list rates (\$0.038), the 9B model recovers its investment after roughly \textbf{80K} samples and the 27B model after roughly \textbf{324K}. Against the cheaper Kimi-K2.6 the 9B model breaks even at 327K samples and the 27B model does not break even at all, because its marginal cost exceeds that API's per-sample price outright. Figure~\ref{fig:cost_breakeven} plots both. We regard 324K as a volume a sustained data-production effort reaches -- it is roughly ten times our own corpus -- but it is not a small number, and a project that will only ever need tens of thousands of samples should call the API instead.

\paragraph{Sensitivity.} A single point estimate would be fragile, because both accelerator rental and API pricing span wide ranges. We therefore recompute $n^{\star}$ for both backbones while varying the three quantities the estimate is most sensitive to: the accelerator rate, which spans nearly $8\times$ between marketplace floors and hyperscaler list prices for the same A100 silicon; the serving throughput; and the API tier, since batch pricing halves both the up-front teacher cost and the per-sample price we compete against. Table~\ref{tab:cost_sensitivity} reports the result.

\begin{table}[t]
\centering
\small
\caption{Break-even volume against Gemini-3.1-Pro under varied assumptions, for both backbones. ``never'' means the marginal cost of local inference exceeds the API price outright, so no production volume recovers the investment. The 9B configuration is robust across most of the range; the 27B configuration is not, and pays off only on competitively priced accelerators at list-tier API prices.}
\label{tab:cost_sensitivity}
\begin{tabular}{lccrr}
\toprule
Assumption & \$/GPU-h & API tier & $n^{\star}$ (9B) & $n^{\star}$ (27B) \\
\midrule
Headline (Table~\ref{tab:cost_assumptions}) & \$1.80 & list & 80K & 324K \\
Marketplace accelerators & \$0.65 & list & 54K & 84K \\
Hyperscaler accelerators & \$5.07 & list & 277K & never \\
Serving throughput 2$\times$ worse & \$1.80 & list & 116K & never \\
Batch-tier API pricing & \$1.80 & batch & 149K & never \\
Batch tier + hyperscaler & \$5.07 & batch & never & never \\
Batch tier + hyperscaler + low throughput & \$5.07 & batch & never & never \\
\bottomrule
\end{tabular}
\end{table}

\paragraph{Reading the result.} Two things follow, and they differ by backbone. For the 9B configuration the qualitative claim is robust over the settings a typical research group operates in: at marketplace or median accelerator prices the investment is recovered between 54K and 149K produced samples. For the 27B configuration the claim is conditional in a way we want to state plainly: it pays off at median or marketplace accelerator prices under list-tier API billing, and does not pay off at all once accelerators are rented at hyperscaler list prices or the API is billed at the batch tier. A practitioner in the latter situation should call the API directly, or use the 9B configuration and accept the accuracy cost.

What the analysis supports is therefore narrower than ``self-hosting is cheaper''. It is that self-hosting is cheaper for repeated production -- new domains, new intents, new schema revisions over the lifetime of a project -- on competitively priced accelerators. One caveat on the quality axis: at 27B the joint model exceeds the teacher on $\text{DataClaw}_0$-val, so the cost comparison understates its advantage; at 9B it trails the teacher on open-ended semantics, so the comparison overstates it. Figure~\ref{fig:cost_breakeven} should be read together with Tables~\ref{tab:main_results} and~\ref{tab:downstream_results} rather than on its own.

\paragraph{Non-monetary considerations.} Two further differences are not captured by Eq.~\ref{eq:breakeven}. Self-hosted tailoring keeps raw streams on-premise, which matters for GUI logs and robot trajectories that may contain screen text, credentials, or workspace imagery that cannot be sent to a third party. It also removes rate limits and provider dependence, which become binding constraints when data production is run continuously at scale.

\section{Limitations and Future Work}
\label{app:limitations}

\paragraph{Dependence on a proprietary teacher.} The tailoring capability is bootstrapped by distilling Gemini-3.1-Pro: the SFT corpus is synthesised by the teacher, and the GRPO stage refines a policy initialised from it. Consequently we do not claim that $\text{DataClaw}_0$ creates capability that the teacher lacks. What we do claim is narrower and, we argue, still practically important: the capability can be amortised into a 9B open-weight model whose marginal cost per sample is that of local inference (Appendix~\ref{app:cost}). The transfer results in Sec.~\ref{subsec:scaling_and_diversity} do not change this: transfer between domains is transfer of structure the teacher already supplied, so it enlarges what a fixed teacher budget buys rather than creating capability the teacher lacked. This framing has a direct consequence for the reward design. Our rewards score formatting, anchor consistency, and conciseness -- properties that are cheap to verify -- but they do not directly measure whether a generated question is informative or whether a reasoning trace is semantically correct. Errors present in the teacher's output, or in the extracted anchors, can therefore be absorbed during SFT and are not necessarily corrected by GRPO. Relaxing this dependence requires a training signal that does not originate from the teacher. Two sources are available in principle: existing human-annotated corpora, used as verifiers rather than as imitation targets, and execution feedback from simulators, which can adjudicate counterfactual or corrective examples that never occur in the recorded stream. We regard this as the most important direction for future work.

\paragraph{Evaluation designed alongside the method.} $\text{DataClaw}_0$-val shares its output schema, anchor definitions, and intent taxonomy with our construction pipeline. It therefore measures compatibility with our formulation of the task, and should not be read as a neutral measure of general data-refinement ability. We mitigate this by treating downstream post-training as the primary criterion, but we have not established that the $\text{DataClaw}_0$-val score is predictive of downstream gain, nor have we validated the weights of the hierarchical scoring function against human judgement. A correlation study across checkpoints and a protocol-complete human evaluation would both be needed before the benchmark could be used as a standalone target.

\paragraph{Single-reference sequence matching.} Our sequence metric and the anchor reward compare a generated action sequence against one reference trajectory. In GUI and embodied settings, several action orderings often achieve the same goal, so a valid alternative path can be scored as an error. The reported sequence numbers should thus be read as a lower bound on tailoring quality. A goal-conditioned criterion, or a reference set admitting multiple valid orderings, would give a fairer measure.

\paragraph{Scope of the empirical study.} Results span three backbone sizes (4B, 9B, 27B), which is enough to locate the capacity threshold between 9B and 27B but not to characterise its shape; a denser sweep, or one extending past 27B, could show whether the joint-training advantage keeps growing or saturates. We also vary capacity only through dense backbones from a single model family, so we cannot say whether the threshold is a function of parameter count, of pretraining data, or of the mixture-of-experts architecture Qwen3.5 uses at larger sizes. Downstream validation covers three tasks, and the video-generation setting in particular uses a small fine-tuning set, so the evidence does not yet establish that the gains transfer across model families and datasets. Finally, coverage is limited to five domains and a fixed schema family, and robustness to intents whose requested schema differs substantially from those seen in training remains untested.

\paragraph{Reproducibility and data provenance.} Several implementation details -- reward coefficients, router configuration, and baseline prompting -- are specified in this appendix but not yet released as code; we intend to release the construction pipeline and evaluation harness. The raw streams we process include egocentric video and GUI interaction logs, which may contain personally identifying content such as faces, screen text, and account identifiers. We use only sources released for research use under their original licences and do not redistribute raw streams, but we did not perform an independent privacy audit, and deployments on newly collected streams should treat consent and de-identification as a prerequisite.

\paragraph{From refinement toward autonomous construction.} $\text{DataClaw}_0$ operates on user-provided raw streams: it refines, filters, and annotates existing trajectories or interaction records rather than creating data from intent alone. A natural extension is to let the model enter embodied, game, or GUI simulation environments, actively explore, and construct supervision from its own interactions. Combined with the execution-grounded rewards discussed above, this would move the system from raw-stream refinement toward self-improving data generation, in which each round consumes new streams and the improvement signal comes from verification rather than imitation.


\clearpage

\end{document}